\documentclass[11pt,a4paper]{article}
\usepackage[hyperref]{emnlp2020}
\usepackage{times}
\usepackage{latexsym}

\usepackage{microtype}

\usepackage[utf8]{inputenc} % allow utf-8 input
\usepackage[T1]{fontenc}    % use 8-bit T1 fonts
\usepackage{hyperref}       % hyperlinks
\usepackage{url}            % simple URL typesetting
\usepackage{booktabs}       % professional-quality tables
\usepackage{amsfonts}       % blackboard math symbols
\usepackage{nicefrac}       % compact symbols for 1/2, etc.
\usepackage{microtype}      % microtypography
\usepackage{amsmath}
\usepackage{mathabx}
\usepackage[pdftex]{graphicx}
\usepackage{subcaption}
\graphicspath{{./images/}}
\usepackage{subcaption}
\usepackage[titlenumbered,boxed,ruled]{algorithm2e}

\aclfinalcopy % Uncomment this line for the final submission
\def\aclpaperid{658} %  Enter the acl Paper ID here 

\title{{D}yERNIE: {D}ynamic {E}volution of {R}iemannian {M}anifold {E}mbeddings for {T}emporal {K}nowledge {G}raph {C}ompletion}

\author{Zhen Han$^{1,2}$, \; Yunpu Ma$^{1,2}$\thanks{\; Corresponding author.} , \; Peng Chen$^{2,3}$, \;  Volker Tresp$^{1,2*}$ \\
$^{1}$Institute of Informatics, LMU Munich $\;$  $^{2}$ Corporate Technology, Siemens AG\\
$^{3}$Department of Informatics, Technical University of Munich\\
\texttt{zhen.han@campus.lmu.de,  peng.chen@tum.de}\\
\texttt{cognitive.yunpu@gmail.com, volker.tresp@siemens.com}\\}
\date{Sep. 21, 2020}
 
\begin{document}
\maketitle

\begin{abstract}
There has recently been increasing interest in learning representations of temporal knowledge graphs (KGs), which record the dynamic relationships between entities over time.  Temporal KGs often exhibit multiple simultaneous non-Euclidean structures, such as hierarchical and cyclic structures. However, existing embedding approaches for temporal KGs typically learn entity representations and their dynamic evolution in the Euclidean space, which might not capture such intrinsic structures very well. To this end, we propose DyERNIE, a non-Euclidean embedding approach that learns evolving entity representations in a product of Riemannian manifolds, where the composed spaces are estimated from the sectional curvatures of underlying data. Product manifolds enable our approach to better reflect a wide variety of geometric structures on temporal KGs. Besides, to capture the evolutionary dynamics of temporal KGs, we let the entity representations evolve according to a velocity vector defined in the tangent space at each timestamp. We analyze in detail the contribution of geometric spaces to representation learning of temporal KGs and evaluate our model on temporal knowledge graph completion tasks. Extensive experiments on three real-world datasets demonstrate significantly improved performance, indicating that the dynamics of multi-relational graph data can be more properly modeled by the evolution of embeddings on Riemannian manifolds. 

\end{abstract}

\section{Introduction}
Learning from relational data has long been considered as a key challenge in artificial intelligence. In recent years, several sizable knowledge graphs (KGs), e.g. Freebase \citep{bollacker2008freebase} and Wikidata \citep{vrandevcic2014wikidata}, have been developed that provide widespread availability of such data and enabled improvements to a plethora of downstream applications such as recommender systems \citep{hildebrandt2019recommender} and 
question answering \citep{zhang2018variational}. KGs are multi-relational, directed graphs with labeled edges, where each edge corresponds to a fact and can be represented as a triple, such as (John, lives in, Vancouver). Common knowledge graphs are static and store facts at their current state. In reality, however, multi-relational data are often time-dependent. For example, the political relationship between two countries might intensify because of trade fights. Thus, temporal knowledge graphs were introduced, such as ICEWS \citep{boschee2015icews} and GDELT \citep{leetaru2013gdelt}, that capture temporal aspects of facts in addition to their multi-relational nature. In these datasets, temporal facts are represented as a quadruple by extending the static triplet with a timestamp describing when these facts occurred, i.e. (Barack Obama, inaugurated, as president of the US, 2009). Since real-world temporal KGs are usually incomplete, the task of link prediction on temporal KGs has gained growing interest. The task is to infer missing facts at specific time-stamps based on the existing ones by answering queries such as (US, president, ?, 2015).

Many facts in temporal knowledge graphs induce geometric structures over time. For instance, increasing trade exchanges and economic cooperation between two major economies might promote the trade exports and economic growths of a series of countries in the downstream supply chain, which exhibits a tree-like structure over time. Moreover, an establishment of diplomatic relations between two countries might lead to regular official visits between these two countries, which produces a cyclic structure over time.  Embedding methods in Euclidean space have limitations and suffer from large distortion when representing large-scale hierarchical data. Recently, hyperbolic geometry has been exploited in several works \citep{nickel2017poincare, ganea2018hyperbolic} as an effective method for learning representations of hierarchical data, where the exponential growth of distance on the boundary of the hyperbolic space naturally allows representing hierarchical structures in a compact form. While most graph-structured data has a wide variety of inherent geometric structures, e.g. partially tree-like and partially cyclical, the above studies model the latent structures in a single geometry with a constant curvature, limiting the flexibility of the model to match the hypothetical intrinsic manifold. Thus, using a product of different constant curvature spaces \citep{gu2018learning} might be helpful to match the underlying geometries of temporal knowledge graphs and provide high-quality representations.

Existing non-Euclidean approaches for knowledge graph embeddings \citep{balazevic2019multi, kolyvakis2019hyperkg} lack the ability to capture temporal dynamics available in underlying data represented by temporal KGs. The difficulty with representing the evolution of temporal KGs in non-Euclidean spaces lies in finding a way to integrate temporal information 
to the geometric representations of entities. In this work, we propose the \underline{\textbf{dy}}namic \underline{\textbf{e}}volution of \underline{\textbf{R}}iemannian ma\underline{\textbf{ni}}fold \underline{\textbf{e}}mbeddings (DyERNIE), a theoretically founded approach to embed multi-relational data with dynamic relationships on a product of Riemannian manifolds with different curvatures. To capture both the stationary and dynamic characteristics of temporal KGs, we characterize the time-dependent representation of an entity as movements on manifolds. For each entity, we define an initial embedding (at $t_0$) on each manifold and a velocity vector residing in the tangent space of the initial embedding to generate a temporal representation at each timestamp. In particular, the initial embeddings represent the stationary structural dependencies across facts, while the velocity vectors capture the time-varying properties of entities.

Our contributions are the following: (i) We introduce Riemannian manifolds as embedding spaces to capture geometric features of temporal KGs. (ii) We characterize the dynamics of temporal KGs as movements of entity embeddings on Riemannian manifolds guided by velocity vectors defined in the tangent space. (iii) We show how the product space can be approximately identified from sectional curvatures of temporal KGs and how to choose the dimensionality of component spaces as well as their curvatures accordingly. (iv) Our approach significantly outperforms current benchmarks on a link prediction task on temporal KGs in low- and high-dimensional settings. (v) We analyze our model's properties, i.e. the influence of embedding dimensionality and the correlation between node degrees and the norm of velocity vectors.

\section{Preliminaries}
\subsection{Riemannian Manifold}
An $n$-dimensional Riemannian \textit{manifold} $\mathcal M^n$ is a real and smooth manifold with locally Euclidean structure.  For each point $\mathbf x \in \mathcal M^n$, the metric tensor $g(\mathbf x)$ defines a positive-definite inner product $g(\mathbf x) = \left \langle \cdot, \cdot \right \rangle _{\mathbf x} : \mathcal T_{\mathbf x} \mathcal M^n \times \mathcal T_{\mathbf x} \mathcal M^n  \rightarrow \mathbb R$, where $ \mathcal T_{\mathbf x} \mathcal M^n$ is the tangent space of $\mathcal M^n$ at $\mathbf x$. From the tangent space $ \mathcal T_{\mathbf x} \mathcal M^n$, there exists a mapping function $\exp_{\mathbf x}(\mathbf v): \mathcal T_{\mathbf x} \mathcal M^n \rightarrow \mathcal M^n$ that maps a tangent vector $\mathbf v$ at $\mathbf x$ to the manifold,  also known as the \textit{exponential map}. The inverse of an exponential map is referred to as the 
\textit{logarithm map} $\log_{\mathbf x}(\cdot)$. 

\subsection{Constant Curvature Spaces}
The sectional curvature $K(\tau_{\mathbf x})$ is a fine-grained notion defined over a two-dimensional subspace $\tau_{\mathbf x}$ in the tangent space at the point $\mathbf x$ \citep{berger2012panoramic}. If all the sectional curvatures in a manifold $\mathcal M^n$ are equal, the manifold then defined as a space with a constant curvature $K$. 
Three different types of constant curvature spaces can be defined depending on the sign of the curvature: a positively curved space, a flat space, and a negatively curved space. There are different models for each constant curvature space. To unify different models, in this work, we choose the stereographically projected hypersphere $\mathbb S^n_K$ for positive curvatures ($K > 0$), while for negative curvatures ($K < 0$) we choose the Poincar\'e ball $\mathbb P^n_K$, which is the stereographic projection of the hyperboloid model:
\begin{equation*}
\mathcal M^n_K = \left\{  
           \begin{aligned}
		  \mathbb S^n_{K} & =  \{\mathbf x \in \mathbb R^n: \left \langle \mathbf x, \mathbf x \right \rangle_2 > -1/K \} \\ %, \; K > 0 \\
		  \mathbb E^n & =  \mathbb R^n, \; \text{if} \, K = 0 \\
	       \mathbb P^n_{K} & =  \{\mathbf x \in \mathbb R^n: \left \langle \mathbf x, \mathbf x \right \rangle_2 < -1/K \} %, \; K < 0.
\end{aligned}
\right.  
\end{equation*}
Both of the above spaces $\mathbb S_{K}$ and $\mathbb P_{K}$ are equipped with the Riemannian metric: $g^{\mathbb S_K}_{\mathbf x} = g^{\mathbb P_K}_{\mathbf x} = (\lambda^K_{\mathbf x})^2 g^{\mathbb E}$, which is conformal to the Euclidean metric $g^{\mathbb E}$ with the conformal factor $\lambda^K_{\mathbf x} = 2/(1 + K||\mathbf x||^2_2)$ \citep{ganea2018hyperbolic}. As explained in \citep{skopek2019mixed}, $\mathbb S_{K}$ and $\mathbb P_{K}$  have a suitable property, namely the distance and the metric tensors of these spaces converge to their Euclidean counterpart as the curvature goes to $0$, which makes both spaces suitable for learning sign-agnostic curvatures. 

\subsection{Gyrovector Spaces} 
An important analogy to vector spaces (vector addition and scalar multiplication) in non-Euclidean geometry is the notion of gyrovector spaces \citep{ungar2008gyrovector}. Both the projected hypersphere and the Poincar\'e ball  share the following definition of \textit{Möbius addition}:
\begin{equation*}
\begin{aligned}
&\mathbf x \oplus _K \mathbf y  = \\
 & \frac{(1 - 2K \left \langle \mathbf x, \mathbf y \right \rangle_2 - K || \mathbf  y||^2_2) \mathbf x + (1 + K || \mathbf  x||^2_2)\mathbf y}{1 - 2 K \left \langle \mathbf x, \mathbf y \right \rangle_2 + K^2 || \mathbf  x||_2^2 || \mathbf  y||_2^2}
 \end{aligned}
\end{equation*}
where we denote the Euclidean norm and inner product by $|| \cdot ||$ and $\left \langle \cdot, \cdot \right \rangle_2$, respectively.  \citet{skopek2019mixed} show that the distance between two points in $\mathbb S_K$ or $\mathbb P_K$ is equivalent to their variants in gyrovector spaces, which is defined as
\begin{equation*}
d_{\mathcal M_K}(\mathbf x, \mathbf y) 
= \frac{2}{\sqrt{|K|}}\tan_K^{-1}(\sqrt{|K|}||-\mathbf x \oplus_K \mathbf y||_2), 
\end{equation*}
where $\tan_K = \tan$ if $K > 0$ and $\tan_K = \tanh$ if $K<0$. The same gyrovector spaces can be used to define the exponential and logarithmic maps in the Poincar\'e ball and the projected hypersphere. We list these mapping functions in Table \ref{tab: Exponential and logarithmic maps} in the appendix. As \citet{ganea2018hyperbolic} use the exponential and logarithmic maps to obtain the \textit{Möbius matrix-vector multiplication}: $ \mathbf M \otimes_K \mathbf x = \exp^K_{\mathbf 0} (\mathbf M \log^K _{\mathbf 0}(\mathbf x))$, 
we reuse them in hyperbolic space.  This operation is defined similarly in projected hyperspherical space. 
 
\subsection{Product Manifold}
We further generalize the embedding space of latent representations from a single manifold to a product of Riemannian manifolds with constant curvatures. Consider a sequence of Riemannian manifolds with constant curvatures, the product manifold is defined as the Cartesian product of $k$ component manifolds $\mathcal M^n = \boldsymbol \bigtimes_{i = 1}^{k} \mathcal M_{K_i}^{n_i}$, where $n_i$ is the dimensionality of the $i-$th component, and $K_i$ indicates its curvature, with choices $\mathcal M_{K_i}^{n_i} \in \{\mathbb P_{K_i}^{n_i}, \mathbb E^{n_i}, \mathbb S_{K_i}^{n_i}\}$. We call $\{(n_i, k_i)\}_{i=1}^{k}$ the \textit{signature} of a product manifold. Note that the notation $\mathbb E^{n_i}$ is redundant in Euclidean spaces since the Cartesian product of Euclidean spaces with different dimensions can be combined into a single space, i.e. $\mathbb E^n =\bigtimes^k_{i =1} \mathbb E^{n_i}$. However, this equality does not hold in the projected hypersphere and the Poincar\'e ball. For each point $\mathbf x \in \mathcal M^n$ on a product manifold, we decompose its coordinates into the corresponding coordinates in component manifolds $\mathbf x =(\mathbf x^{(1)}, ..., \mathbf x^{(k)})$, where $\mathbf x^{(i)} \in \mathcal M_{K_i}^{n_i}$. The distance function decomposes based on its definition $d_{\mathcal M^n}^2(\mathbf x, \mathbf y) = \sum_{i=1}^k d_{\mathcal M^{n_i}_{K_i}}^2(\mathbf x^{(i)}, \mathbf y^{(i)})$. Similarly, we decompose the metric tensor, exponential and logarithmic maps on a product manifold  into the component manifolds. In particular, we split the embedding vectors into parts $\mathbf x^{(i)}$, apply the desired operation on that part $f^{n_i}_{K_i}(\mathbf x^{(i)})$, and concatenate the resulting parts back \citep{skopek2019mixed}.

\subsection{Temporal Knowledge Graph Completion}
Temporal knowledge graphs (KGs) are multi-relational, directed graphs with labeled timestamped edges between entities. Let $\mathcal E$, $\mathcal P$, and $\mathcal T$ represent a finite set of entities, predicates, and timestamps, respectively. Each fact can be denoted by a quadruple $q = (e_s, p, e_o, t)$, where $p \in \mathcal P$ represents a timestamped and labeled edge between a subject entity $e_s \in \mathcal E$ and an object entity $e_o \in \mathcal E$ at a timestamp $t \in \mathcal T$.  Let $\mathcal F$ represents the set of all quadruples that are facts, i.e. real events in the world, the temporal knowledge graph completion (tKGC) is the problem of inferring $\mathcal F$ based on a set of observed facts $\mathcal O$, which is a subset of $\mathcal F$. To evaluate the proposed algorithms, the task of tKGC is to predict either a missing subject entity $(?, p, e_o, t)$ given the other three components or a missing object entity $(e_s, p, ?, t)$. Taking the object prediction as an example, we consider all entities in the set $\mathcal E$, and learn a score function $\phi: \mathcal E \times \mathcal P \times \mathcal E \times \mathcal T \rightarrow \mathbb R$.  Since the score function assigns a score to each quadruple, the proper object can be inferred by ranking the scores of all quadruples $\{(e_s, p, e_{o_i}, t),e_{o_i} \in \mathcal E \}$ that are accompanied with candidate entities. 

\section{Related work}
\subsection{Knowledge Graph Embedding}
\paragraph{Static KG Embedding}
Embedding approaches for static KGs can generally be categorized into bilinear models and translational models. 
The bilinear approaches are equipped with a bilinear score function that represents predicates as linear transformations acting on entity embeddings \citep{ nickel2011three, trouillon2016complex, yang2014embedding, ma2018holistic}. Translational approaches measure the plausibility of a triple as the distance between the translated subject and object entity embeddings, including TransE \citep{bordes2013translating} and its variations \citep{sun2019rotate, kazemi2018simple}. Additionally, several models are based on deep learning approaches \citep{nguyen2017novel, dettmers2018convolutional, schlichtkrull2018modeling, nathani2019learning, hildebrandt2020reasoning} that apply (graph) convolutional layers on top of embeddings and design a score function as the last layer of the neural network.

\paragraph{Temporal KG Embedding}
Recently, there have been some attempts of incorporating time information in temporal KGs to improve the performance of link prediction. 
\citet{ma2018embedding} developed extensions of static knowledge graph models by adding a timestamp embedding to the score functions. Also, \citet{leblay2018deriving} proposed TTransE by incorporating time representations into the score function of TransE in different ways. 
HyTE \citep{dasgupta2018hyte} embeds time information in the entity-relation space by arranging a temporal hyperplane to each timestamp. 
Inspired by the canonical decomposition of tensors, \citet{lacroix2020tensor} proposed an extension of ComplEx \citep{trouillon2016complex} for temporal KG completion by decomposing a 4-way tensor. The number of parameters of these models scales with the number of timestamps, leading to overfitting when the number of timestamps is extremely large.  Additionally, a considerable amount of models \citep{trivedi2017know, jin2019recurrent, han2020graph} have been developed for forecasting on temporal knowledge graphs, which predict future links only based on past events.

\subsection{Graph Embedding Approaches in non-Euclidean Geometries}
There has been a growing interest in embedding graph data in non-Euclidean spaces.        \citet{nickel2017poincare} first applied hyperbolic embedding for link prediction to the lexical database WordNet. Since then, hyperbolic analogs of several other approaches have been developed \citep{de2018representation, tifrea2018poincar}. 
In particular, 
\citet{balazevic2019multi} proposed a translational model for embedding multi-relational graph data in the hyperbolic space and demonstrated advancements over state-over-the-art. More recently, \citet{gu2018learning} generalized manifolds of constant curvature to a product manifold combining hyperbolic, spherical, and Euclidean components. However, these methods consider graph data as static models and lack the ability to capture temporally evolving dynamics.

\section{Temporal Knowledge Graph Completion in Riemannian Manifold}
Entities in a temporal KG might form different geometric structures under different relations, and these structures could evolve with time. 
To capture heterogeneous and time-dependent structures, we propose the DyERNIE model to embed entities of temporal knowledge graphs on a product of Riemannian manifolds and model time-dependent behavior of entities with dynamic entity representations. 

\subsection{Entity Representation} 
\begin{figure}
\begin{subfigure}{.22\textwidth}
 \centering
   \includegraphics[width=\linewidth]{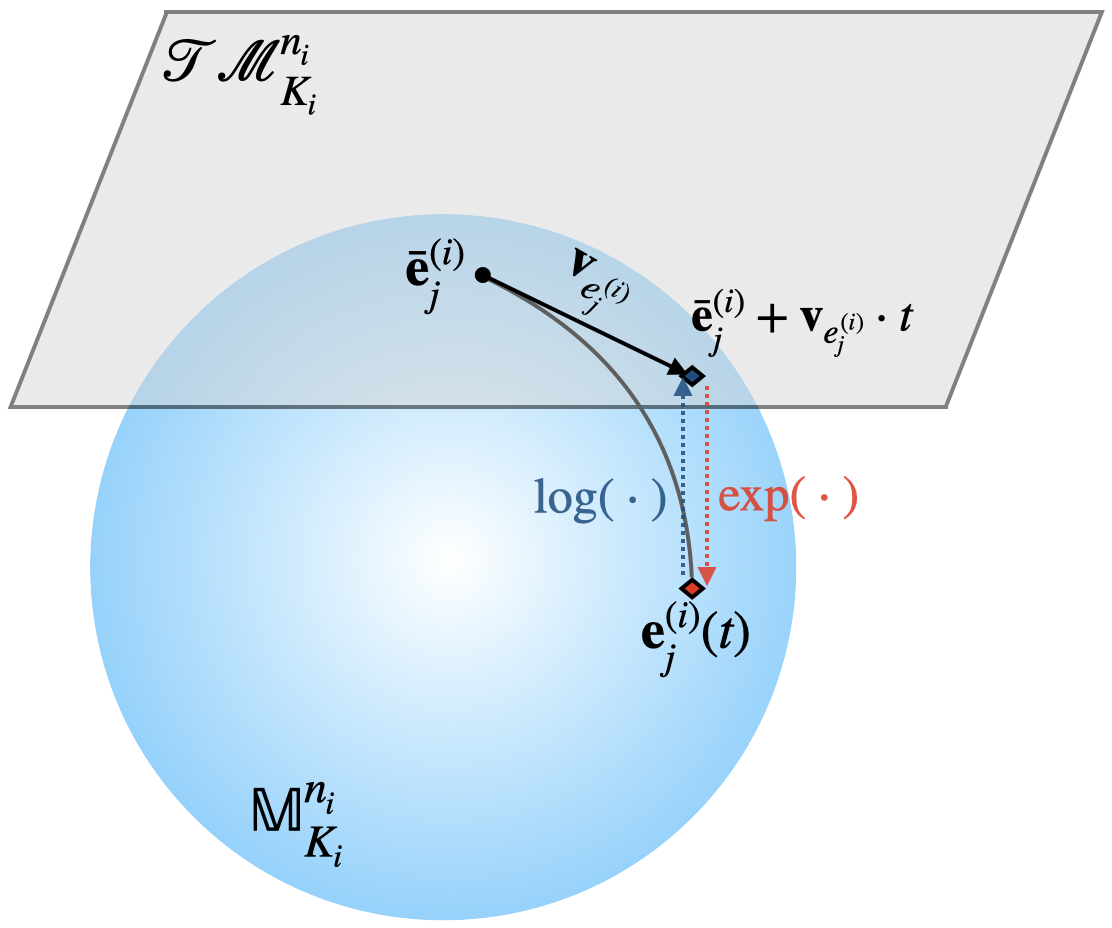}
\end{subfigure}%
\begin{subfigure}{.22\textwidth}
 \centering
  \includegraphics[width=.7\linewidth]{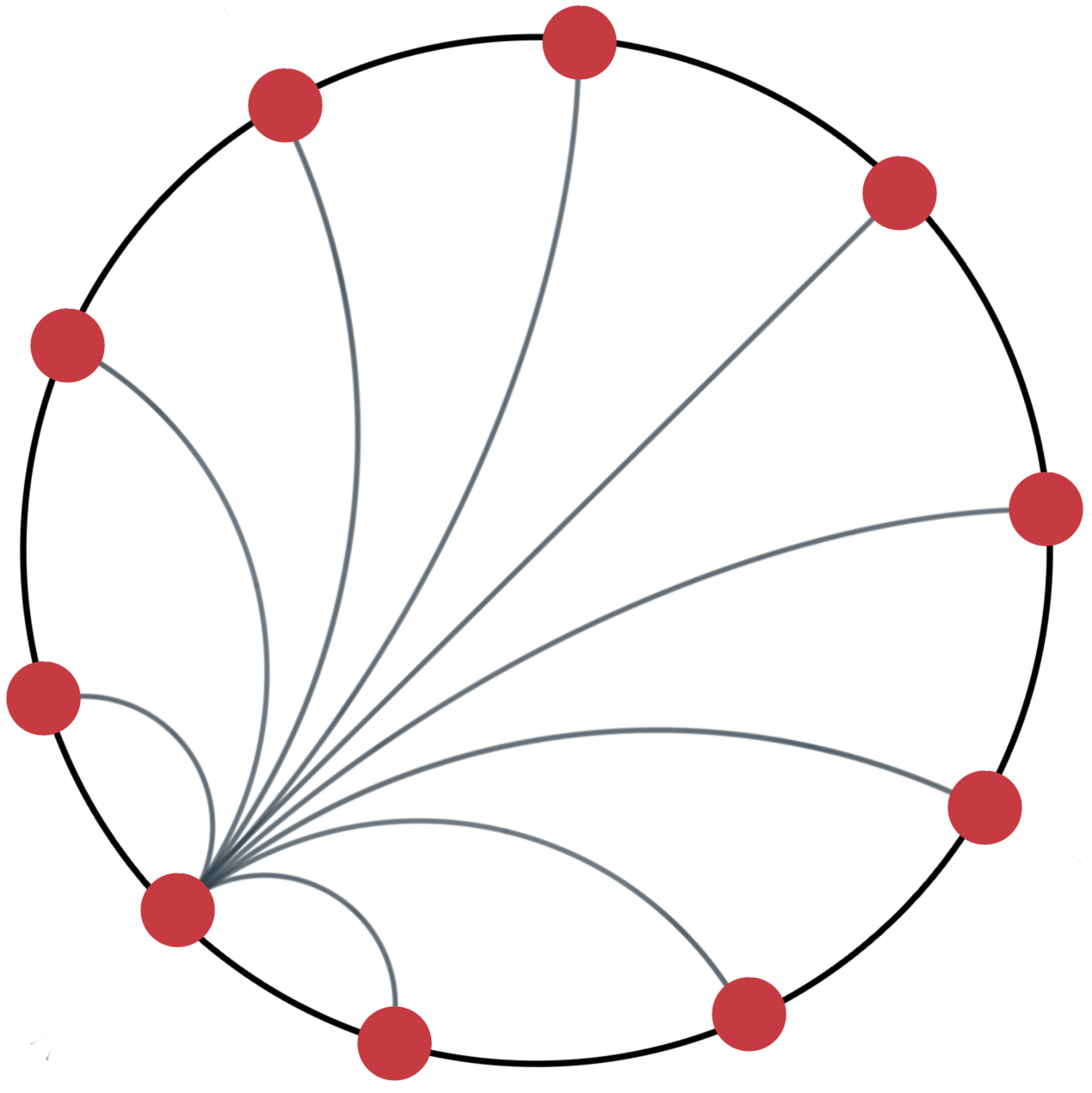}
\end{subfigure}
\caption{\label{fig:Velocity vector of an entity embedding}(a) Evolution of an entity embedding on the $i$-th component manifold (left). For convenience in drawing, the tangent space $\mathcal T \mathcal M^{n_i}_{K_i}$ is defined at $\mathbf{\bar e}_{j}^{(i)}$. (b) Geodesics in the Poincar\'e  disk (right), where red dots represent nodes on the disk.}
\end{figure}

In temporal knowledge graphs, entities might have some features that change over time and some features that remain fixed. Thus, we represent the embedding of an entity $e_j \in \mathcal E$ at instance $t$ with a combination of low-dimensional vectors $\mathbf e_j(t) = (\mathbf e_j^{(1)}(t), ..., \mathbf e_j^{(k)}(t))$ with $ \mathbf e_j^{(i)}(t) \in \mathcal M^{n_i}_{K_i}$, where $\mathcal M^{n_i}_{K_i} \in \{\mathbb P_{K_i}^{n_i}, \mathbb E^{n_i}, \mathbb S_{K_i}^{n_i}\}$ is the $i$-th component manifold, $K_i$ and $n_i$ denote the curvature and the dimension of this manifold, respectively. Each component embedding $\mathbf e_j^{(i)}(t)$ is derived from an initial embedding and a velocity vector to encode both the stationary properties of the entities and their time-varying behavior, namely 
\begin{equation}
\label{equa: entity representation}
\mathbf e_{j}^{(i)}(t) = \exp^{K_i}_{\mathbf 0}\left(\log^{K_i}_{\mathbf 0}(\mathbf{\bar e}_j^{(i)}) + \mathbf v_{e_j^{(i)}}t\right),
\end{equation}
where $\mathbf{\bar e}_j^{(i)} \in \mathcal M^{n_i}_{K_i}$ represents the initial embedding that does not change over time. $\mathbf v_{e_j^{(i)}} \in \mathcal T_{\mathbf 0}{\mathcal M^{n_i}_{K_i}}$ represents an entity-specific velocity vector that is defined in the tangent space at origin $\mathbf 0$ and captures evolutionary dynamics of the entity $e_j$ in its vector space representations over time. As shown in Figure \ref{fig:Velocity vector of an entity embedding} (a), we project the initial embedding to the tangent space $\mathcal T_{\mathbf 0}{\mathcal M^{n_i}_{K_i}}$ using the logarithmic map $\log^{K_i}_{\mathbf 0}$ and then use a velocity vector to obtain the embedding of the next timestamp. Finally, we project it back to the manifold with the exponential map $\exp^{K_i}_{\mathbf 0}$. Note that in the case of Euclidean space, the exponential map and the logarithmic map are equal to the identity function. By learning both the initial embedding and velocity vector, our model 
characterizes evolutionary dynamics of entities as movements on manifolds and thus predict unseen entity interactions based on both the stationary and time-varying entity properties.

\subsection{Score Function}  
Bilinear models have been proved to be an effective approach for KG completion \citep{nickel2011three, lacroix2018canonical}, where the score function is a bilinear product between subject entity, predicate, and object entity embeddings. However, there is no clear correspondence of the Euclidean inner-product in non-Euclidean spaces. We follow the method suggested in Poincar\' e Glove \citep{tifrea2018poincar} to reformulate the inner product as a function of distance, i.e. $\left \langle \mathbf x, \mathbf y \right \rangle = \frac{1}{2}(d(\mathbf x, \mathbf y)^2 + ||x||^2 + ||y||^2)$ and replace squared norms with biases $b_{\mathbf x}$ and $b_{\mathbf y}$. In addition, to capture different hierarchical structures under different relations simultaneously,  \citet{balazevic2019multi} applied relation-specific transformations to entities, i.e. a stretch by a diagonal predicate matrix $\mathbf P \in \mathbb R^{n \times n}$ to subject entities and a translation by a vector offset $\mathbf p \in \mathbb P^n$ to object entities.

Inspired by these two ideas, we define the score function of DyERNIE as
\begin{equation*}
\label{equa: score function}
\begin{aligned}
& \phi (e_s, p, e_o, t) = \sum_{i=1}^{k} -d_{\mathcal M^{n_i}_{K_i}} \Big(\mathbf P^{(i)} \otimes_{K_i} \mathbf e_{s}^{(i)}(t), \\
&  \mathbf e_{o}^{(i)}(t) \oplus_{K_i} \mathbf p^{(i)}\Big)^2 
+ b_s^{(i)} + b_o^{(i)}
\end{aligned}
\end{equation*}
where $\mathbf e_{s}^{(i)}(t)$ and $\mathbf e_{o}^{(i)} (t) \in \mathcal M^{n_i}_{K_i}$ are  embeddings of the subject and object entities $e_s$ and $e_o$ in the $i$-th component manifold, respectively. $\mathbf p^{(i)} \in \mathcal M^{n_i}_{K_i}$  is a translation vector of predicate $p$, and $\mathbf P^{(i)} \in \mathbb R^{n_i \times n_i}$ represents a diagonal predicate matrix defined in the tangent space at the origin. 
Since multi-relational data often has different structures under different predicate, we use predicate-specific transformations $\mathbf P$ and $\mathbf p$ to determine the predicate-adjusted embeddings of entities in different predicate-dependent structures, e.g. multiple hierarchies. The distance between the predicate-adjusted embeddings of $e_s$ and $e_o$ measures the relatedness between them in terms of a predicate $p$.  

\begin{figure}
\begin{subfigure}{.2\textwidth}
 \centering
   \includegraphics[width=\linewidth]{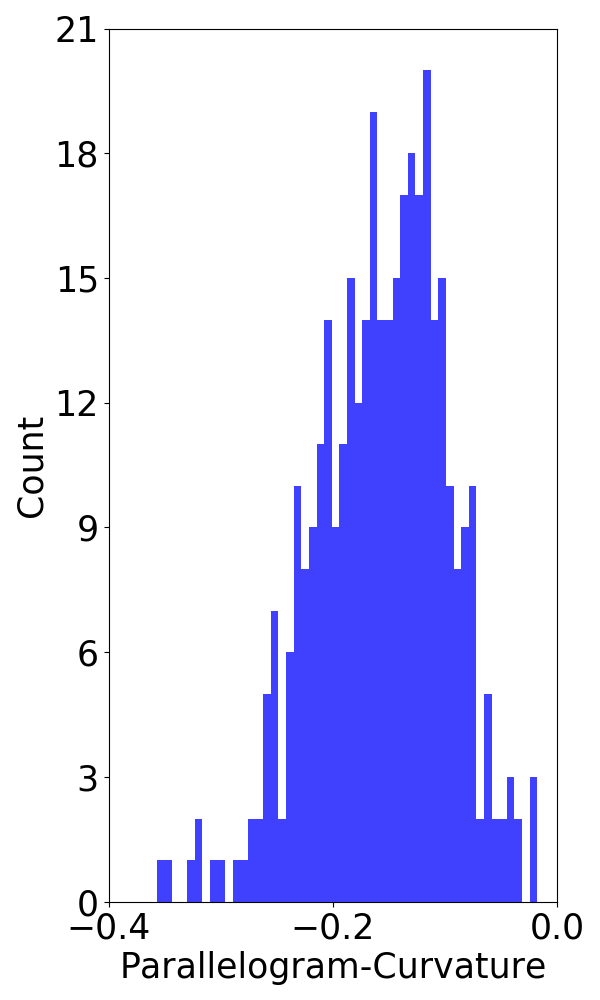}
\end{subfigure}%
\hspace{4mm}
\begin{subfigure}{.2\textwidth}
 \centering
  \includegraphics[width=\linewidth]{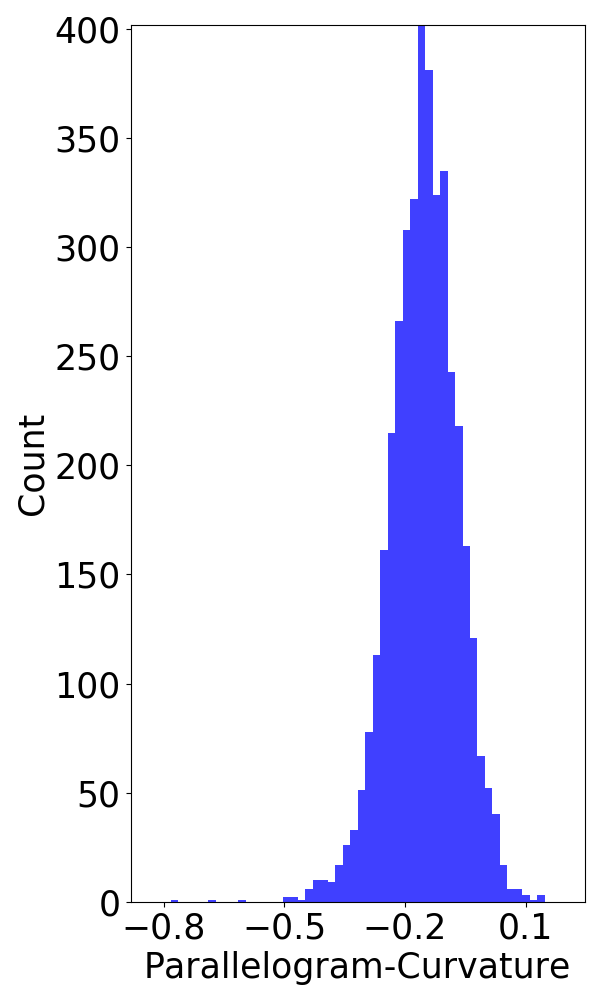}
\end{subfigure}%
\caption{\label{fig:Histogram of time slice curvatures} Histogram of sectional curvatures at each timestamps on ICEWS14 (left), and ICEWS05-15 (right).}
\end{figure}

\subsection{Learning}
The genuine quadruples in a temporal KG $\mathcal G$ are split into \textit{train}, \textit{validation}, and \textit{test} sets. We add reciprocal relations for every quadruple, which is a standard data augmentation technique commonly used in literature \citep{balazevic2019multi, goel2019diachronic}, i.e. we add $(e_o, p^{-1}, e_s, t)$ for every $(e_s, p, e_o, t)$.  Besides, for each fact $(e_s, p, e_o, t)$ in  the training set, we generate $n$ negative samples by corrupting either the object $(e_s, p, e_o', t)$ or the subject $(e_o, p^{-1}, e_s', t)$ with a randomly selected entity from $\mathcal E$. 
We use the binary cross-entropy as the loss function, which is defined as
\begin{equation*}
\begin{aligned}
&\mathcal L = \\
&\frac{-1}{N}\sum_{m=1}^N \left(y_m \log(p_m) + (1-y_m) \log(1-p_m)\right),
\end{aligned}
\end{equation*}
where $N$ is the number of training samples, $y_m$ represents the binary label indicating whether a quadruple $q_m$ is genuine or not, $p_m$ denotes the predicted probability $\sigma(\phi(q_m))$, and $\sigma(\cdot)$ represents the sigmoid function. Model parameters are learned using \textit{Riemannian stochastic gradient descent} (RSGD) \citep{bonnabel2013stochastic}, where the Riemannian gradient $\nabla_{\mathcal M^n} L$ is obtained by multiplying the Euclidean gradient $\nabla_{\mathbb E}$ with the inverse of the Riemannian metric tensor.

\subsection{Signature Estimation}
To better capture a broad range of structures in temporal KGs, we need to choose an appropriate signature of a product manifold $\mathcal M^n$, including the number of component spaces, their dimensions, and curvatures. Although we can simultaneously learn embeddings and the curvature of each component during training using gradient-based optimization, we have empirically found that treating curvature as a trainable parameter interferes with the training of other model parameters. Thus, we treat the curvature of each component and the dimension as hyperparameters selected \textit{a priori}. In particular, we use the parallelogram law' deviation \citep{gu2018learning} to estimate both the graph curvature of a given temporal KG and the number of components. Details about this algorithm can be found in Appendix A. Figure \ref{fig:Histogram of time slice curvatures} shows the curvature histograms on the ICEWS14 and ICEWS05-15 datasets introduced in Section \ref{para: datasets}. It can be noticed that curvatures are mostly non-Euclidean, offering a good motivation to learn embeddings on a product manifold. Taking the ICEWS05-15 dataset as an example, we see that most curvatures are negative. In this case, we can initialize the product manifold consisting of three hyperbolic components with different dimensions. 
Then we conduct a Bayesian optimization around the initial value of the dimension and the curvature of each component to fine-tune them. Finally, we select the best-performing signature according to performance on the validation set as the final choice.

\section{Experiments}
\label{sec: experiments}
\subsection{Experimental Set-up}
\paragraph{\label{para: datasets}Datasets} 
Global Database of Events, Language, and Tone (GDELT) \citep{leetaru2013gdelt} dataset and Integrated Crisis Early Warning System (ICEWS) \citep{boschee2015icews} dataset 
have established themselves in the research community as representative samples of temporal KGs. The GDELT dataset is derived from an initiative database of all the events across the globe connecting people, organizations, and news sources. We use a subset extracted by \citet{jin2019recurrent}, which contains events occurring from 2018-01-01 to 2018-01-31. The ICEWS dataset contains information about political events with specific time annotations, e.g. (Barack Obama, visit, India, 2010-11-06). We apply our model on two subsets of the ICEWS dataset generated by \citet{garcia2018learning}: ICEWS14 contains events in 2014, and ICEWS05-15 corresponds to the facts between 2005 to 2015. 
We compare our approach and baseline methods by performing the link prediction task on the GDELT, ICEWS14 and ICEWS05-15 datasets. The statistics of the datasets are provided in Appendix C.

\paragraph{Baselines}
Our baselines include both static and temporal KG embedding models. From the static
KG embedding models, we use TransE \citep{bordes2013translating}, DistMult \citep{yang2014embedding}, and ComplEx \citep{trouillon2016complex} where we compress temporal knowledge graphs into a static, cumulative graph by ignoring the time information. From the temporal KG embedding models, we compare the performance of our model with several state-of-the-art methods, including TTransE \citep{leblay2018deriving}, TDistMult/TComplEx \citep{ma2018embedding}, 
and HyTE \citep{dasgupta2018hyte}. 

	\begin{table*}[ht!]
    \caption{Link prediction results: MRR (\%) and Hits@1/3/10 (\%). The best results among all models are in bold. Additionally, we underline the best results among models with the same embedding dimension.}
    \label{tab: link prediction results}
    \begin{center}
      \resizebox{\textwidth}{!}{
    \begin{tabular}{l|l|ccccc|ccccc|ccccc} 
      \toprule % <-- Toprule here
     \multicolumn{2}{l}{Datasets} & \multicolumn{5}{|c}{\textbf{ICEWS14 - filtered}} &  \multicolumn{5}{|c}{\textbf{ICEWS05-15 - filtered}} & \multicolumn{5}{|c}{\textbf{GDELT - filtered}}\\
      \midrule % <-- Midrule here
      Rank ($n$) & Model & Manifold & MRR & Hits@1 & Hits@3 & Hits@10 & Manifold & MRR & Hits@1 & Hits@3  & Hits@10 & Manifold & MRR & Hits@1 & Hits@3  & Hits@10 \\
       \midrule % <-- Midrule here
        & TransE &  & 30.0 & 14.8 & 42.7 & 60.1 & & 30.4 & 13.3 & 42.4 & 61.1 &  & 17.7 & 7.9 & 22.9 & 36.8 \\
       100 & DistMult & $\mathbb E$ & 57.5 & 46.9 & 64.2 & 77.9 & $\mathbb E$ & 47.1 & 33.6 & 55.1 & 72.5 & $\mathbb E$ & 22.6 & 13.9 & 26.1 & 39.2\\
        & ComplEx &  & 49.3 & 36.6 & 56.2 & 74.2 & & 39.0 & 22.9 & 49.2 & 68.4 & & 18.8 & 10.5 & 22.2 & 34.9\\
      \midrule % <-- Midrule here
       & TTransE & & 34.4 & 25.7 & 38.3 & 51.3 & &  35.6 & 15.4 & 51.1 & 67.6 & & 18.2 & 0.0 & 30.7 & 46.2 \\
       & TDistMult & & 33.1 & 25.4 & 36.2 & 47.8 & & 49.8 & 41.1 & 54.3 & 66.4 & & 28.3 & 16.2 & 30.7 & 47.1 \\
      100 & TComplEx & $\mathbb E$ & 31.8 & 12.9 & 45.7 & 63.0 & $\mathbb E$ & 45.1 & 36.3 & 49.2 & 62.0 & $\mathbb E$ & 30.6 & 21.0 & 34.7 & 48.1 \\
       & HyTE & & 33.1 & 6.8 & 54.5 & 73.6 &  & 38.1 & 7.6 & 65.0 & 80.4 & & 22.4 & 0.0 & 39.5 & 54.2\\
          \midrule % <-- Midrule here
        & DyERNIE-Prod & $\mathbb P^3$ &  \underline{46.2} & \underline{36.0} & \underline{51.1} & \underline{66.3} & $\mathbb P^3$ & \underline{58.9} & \underline{50.5} & \underline{63.2} & \underline{75.1} & $\mathbb S^2$ & \underline{36.3} & \underline{29.4} & \underline{38.3} & \underline{49.5} \\ 
           10 & DyERNIE-Sgl  & $\mathbb P$ & 43.3 & 33.3 & 47.6 & 62.9 & $\mathbb P$  & 58.0 & 49.2 & 62.8 & 74.5 & $\mathbb S$ & 35.7 & 28.7 & 37.7 & 48.9 \\ 
           & DyERNIE-Euclid & $\mathbb E$ & 39.8 & 30.6 & 43.6 & 58.2 & $\mathbb E$ & 51.9 & 43.4 & 56.1 & 67.9 & $\mathbb E$ & 30.2 & 23.8 & 31.8 & 42.5\\
           \midrule % <-- Midrule here
       & DyERNIE-Prod & $\mathbb P^3$  &  \underline{53.9} &  \underline{44.2} &  \underline{58.9} &  \underline{72.7} & $\mathbb P^3$ &  \underline{64.2} &  \underline{56.5} &  \underline{68.2} &  \underline{79.0} & $\mathbb S^2$ &  \underline{40.0} &  \underline{33.2} & \underline{42.0} & \underline{53.1}\\
           20 & DyERNIE-Sgl  & $\mathbb P$  & 51.3 & 41.4 & 56.1 & 70.3 & $\mathbb P$  & 63.8 & 55.9 & 67.9 & 78.7 & $\mathbb  S$ & 39.2 & 32.6 &  41.1 &  52.1\\  
              & DyERNIE-Euclid  & $\mathbb E$ & 47.7 & 38.3 & 52.0 & 66.2 & $\mathbb E$ & 57.3 & 49.4 & 61.1 & 72.4 & $\mathbb E$ & 32.9 & 26.2 & 34.7 & 45.7\\
     \midrule % <-- Midrule here
         & DyERNIE-Prod &  $\mathbb P^3$  & \underline{58.8} &  \underline{49.8} &  \underline{63.8} &  \underline{76.1} & $\mathbb P^3$ &  \underline{68.9} &  \underline{61.8} &  \underline{72.8} &  \underline{82.5} &  $\mathbb S^2$ &  \underline{43.0} &  \underline{36.3} &  \underline{45.1} &  \underline{56.0}\\
         40 & DyERNIE-Sgl  & $\mathbb P$  & 56.6 & 47.3 & 61.3 &  74.6 &$\mathbb P$  & 67.3 & 60.2 & 71.1 & 81.1 & $\mathbb S $ & 42.5 &  35.8 & 44.6 & 55.6\\
          & DyERNIE-Euclid & $\mathbb E$ & 53.7 & 44.2 & 58.6 & 71.9 & $\mathbb E$ & 60.3 & 52.7 & 64.1 & 74.7& $\mathbb E$ & 38.4 & 31.8 & 40.4 & 51.1\\
           \midrule % <-- Midrule here
         & DyERNIE-Prod & $\mathbb P^3$ &  \textbf{66.9} & \textbf{59.9} & \textbf{71.4} & \textbf{79.7} & $\mathbb P^3$ & \textbf{73.9}  & \textbf{67.9}  & \textbf{77.3} & \textbf{85.5} &  $\mathbb S^2$ & \textbf{45.7} & \textbf{39.0} & \textbf{47.9} & \textbf{58.9} \\       
          100 & DyERNIE-Sgl & $\mathbb P$ &  65.7 & 58.2 & 70.2 & 79.4 &  $\mathbb P$ & 71.2 & 64.8 & 74.6 & 83.4 & $\mathbb S$ & 45.4 & 38.6 & 47.6 & 58.4 \\
        & DyERNIE-Euclid & $\mathbb E$ & 63.3 & 54.9 & 67.9 & 79.2 & $\mathbb E$ & 66.2 & 59.0 & 69.9 & 79.8 & $\mathbb E$ & 42.6 & 36.1 & 44.5 & 55.1\\
      \bottomrule % <-- Bottomrule here
    \end{tabular} }
    \end{center}
	\end{table*}
	
\paragraph{Evaluation protocol}
For each quadruple $q = (e_s, p, e_o, t)$ in the test set $\mathcal G_{test}$, we create two queries: $(e_s, p, ?, t)$ and  $(e_o, p^{-1}, ?, t)$. For each query, the model ranks all possible entities $\mathcal E$ according to their scores. Following the commonly filtered setting in the literature \citep{bordes2013translating}, we remove all entity candidates that correspond to true triples\footnote{The triplets that appear either in the train, validation, or test set.} from the candidate list apart from the current test entity. 
Let $\psi_{e_s}$ and $\psi_{e_o}$ represent the rank for $e_s$ and $e_o$ of the two queries respectively, we evaluate our models using standard metrics across the link prediction literature: \textit{mean reciprocal rank (MRR)}: $\frac{1}{2\cdot |\mathcal G_{test}|} \sum_{q \in \mathcal G_{test}}(\frac{1}{\psi_{e_s}} + \frac{1}{\psi_{e_o}})$ and \textit{Hits}$@k(k \in \{1,3,10\})$: the percentage of times that the true entity candidate appears in the top $k$ of ranked  candidates.

\paragraph{Implementations}
We implemented our model and all baselines in PyTorch \citep{paszke2019pytorch}. For fairness of comparison, we use Table 2 in supplementary materials to compute the embedding dimension for each (baseline, dataset) pair that matches the number of parameters of 
our model with an embedding dimension of $100$. Taking HyTE as an example, its embedding dimension is 193 and 151 on the ICEWS14 and GDELT dataset, respectively.  Also, we use the datasets augmented with reciprocal relations to train all baseline models. We tune hyperparameters of our models using the quasi-random search followed by Bayesian optimization \citep{ruffinelli2020you} and report the best configuration in Appendix E.
We implement TTransE, TComplEx, and TDistMult based on the implementation of TransE, Distmult, and ComplEx respectively. We use the binary cross-entropy loss and RSGD to train these baselines and optimize hyperparameters by early stopping according to MRR on the validation set. Additionally, we use the implementation of HyTE\footnote{https://github.com/malllabiisc/HyTE}. 
We provide the detailed settings of hyperparameters of each baseline model in Appendix B.

\subsection{Comparative Study}
\begin{table}[htbp]
  \begin{center}
    \caption{Filtered MRR for different choices of the distance function with $K=-1$ and $n = 40$ on ICEWS14.}
    \label{tab: distance function variantions}
   \resizebox{0.35\textwidth}{!}{
    \begin{tabular}{lcccc}
      \toprule % <-- Toprule here
  	Distance function & MRR \\
  	 \midrule % <-- Midrule here
  	 $d(\mathbf P \otimes \mathbf e_s (t), \mathbf e_o (t) \oplus \mathbf p)$ & \textbf{55.87} \\ 
  	 	 \midrule % <-- Midrule here
  	 	 $\cosh(d(\mathbf P \otimes \mathbf e_s (t), \mathbf e_o (t) \oplus \mathbf p))$ & 54.00\\
  	   	   	 $d(\mathbf P \otimes \mathbf e_s (t), \mathbf P \otimes \mathbf e_o (t))$ & 52.23  \\ 
  	 $d(\mathbf P \otimes \mathbf e_s (t), \mathbf P \otimes \mathbf e_o (t) \oplus \mathbf p)$ &  54.55  \\ 
  	   $d(\mathbf P \otimes \mathbf e_s (t), \mathbf e_o (t))$ & 47.24 \\
  	    $d(\mathbf e_s (t), \mathbf e_o (t) \oplus \mathbf p)$  & 51.36 \\
      \bottomrule % <-- Bottomrule here
    \end{tabular}
}
  \end{center}
\end{table}
\paragraph{Model variants}
To compare the performance of non-Euclidean embeddings with their Euclidean counterparts, we implement the Euclidean version of Equation \ref{equa: score function} with $d_{\mathcal M}(\mathbf x, \mathbf y) = d_{\mathbb E} (\mathbf x, \mathbf y)$. We refer to it as DyERNIE-Euclid. Besides, we train our model with a single non-Euclidean component to compare embeddings in a product space and in a manifold with a constant curvature. We refer to them as DyERNIE-Prod and DyERNIE-Sgl, respectively. 
For DyERNIE-Prod, we generate model configurations with different manifold combinations, i.e. $\mathbb P \times \mathbb S \times \mathbb E, \mathbb P^3$.  Details about the search space are relegated to Appendix E.

\paragraph{Link prediction results}
We compare the baselines with three variants of our model: DyERNIE-Prod, DyERNIE-Sgl, and DyERNIE-Euclid. We report the best results on the test set among all model configurations in Table \ref{tab: link prediction results}. 
Note that the number of parameters of all baselines matches our model's with an embedding dimension of $100$. Thus, we see that both DyERNIE-Prod and DyERNIE-Sgl significantly outperform the baselines and DyERNIE-Euclid on all three datasets with the same number of parameters. Even at a low embedding dimension $(n = 10)$,  our models still have competitive performance, demonstrating the merits of time-dependent non-Euclidean embeddings. 
Besides, DyERNIE-Prod generally performs better than DyERNIE-Sgl on all three datasets. On the ICEWS14 and ICEWS05-15 datasets, we can observe that the best performing configuration of DyERNIE-Prod at each dimensionality only contains hyperbolic component manifolds. This observation confirms the curvature estimation shown in Figure \ref{fig:Histogram of time slice curvatures}, where most sectional curvatures on the ICEWS14 and ICEWS05-15 datasets are negative. 
\begin{table}[htbp]
  \begin{center}
    \caption{Filtered MRR for different choices of entity representations with $K=-1$ and $n = 40$ on ICEWS14, where $\mathbf A_{i}$ and $\mathbf w_{i}$ represent the amplitude vector and the frequency vector, respectively. $\phi_{i}$ denotes the phase shift.}
    \label{tab: entity representation variantions}
    \resizebox{0.4\textwidth}{!}{
    \begin{tabular}{lcccc}
      \toprule % <-- Toprule here
  	Entity Representations & MRR\\
  	 \midrule % <-- Midrule here
  	$ \exp(\log(\mathbf{\bar e}_i) + \mathbf v_{i}t)$ & \textbf{55.87}\\
  	  \midrule % <-- Midrule here
  	  $ \exp(\log(\mathbf{\bar e}_i) + \mathbf A_{i}\sin(\mathbf w_{i}t + \phi_{i}))$ & 52.50 \\
  	   $ \exp(\log(\mathbf{\bar e}_i) +  \mathbf v_{i}t  + \mathbf A_{i}\sin(\mathbf w_{i}t + \phi_{i})) $ & 53.52\\

      \bottomrule % <-- Bottomrule here
    \end{tabular}
}
  \end{center}
\end{table}
\paragraph{Ablation study}
\label{para: ablation study}
We show an ablation study of the distance function and the entity representations in Table \ref{tab: distance function variantions} and \ref{tab: entity representation variantions}, respectively. For the distance function, we use $\mathbf p$ and  $\mathbf P$ to get predicate-adjusted subject and object embeddings and compute the distance between them. We found that any change to distance function causes performance degradation. Especially, removing the translation vector $\mathbf p$ most strongly decrease the performance. For the entity representation function, we measure the importance of a linear trend component and a non-linear periodic component. We attempt adding trigonometric functions into entity representations since a combination of trigonometric functions can capture more complicated non-linear dynamics \citep{rahimi2008random}. However, experimental results in Table \ref{tab: entity representation variantions} show that using only a linear transformation works the best, which indicates that finding the correct manifold of embedding space is more important than designing complicated non-linear evolution functions of entity embeddings. Additionally, we found the performance degrades significantly if removing the dynamic part of the entity embeddings. For example, on the ICEWS0515 dataset, the Hits@1 metric in the static case is only about half of that in the dynamic case, clearly showing the gain from the dynamism.  Details of this ablation study are provided in Appendix G.

\begin{figure}
  \centering
  \includegraphics[width=.9\linewidth]{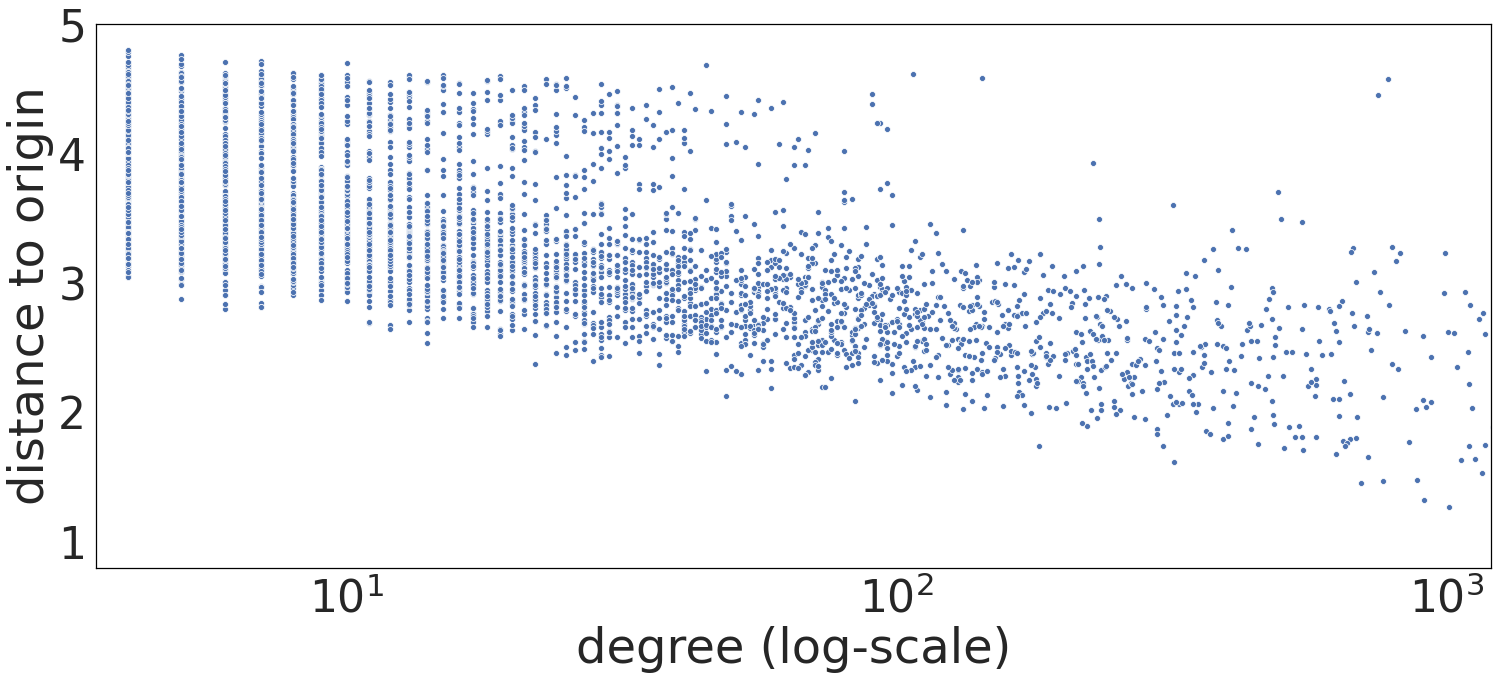}
\caption{\label{fig:degreeAndDistanceOrigin} Scatter plot of distances between entity embeddings and the manifold's origin v.s. node degrees on ICEWS05-15. Each point denotes an entity $e_j$. The x-coordinate gives its degree accumulated over all timestamps, and the y-coordinate represents $d_{\mathcal M}(\mathbf e_j, \mathbf 0)$.}
\end{figure}

\begin{figure}
  \centering
  \includegraphics[width=.9\linewidth]{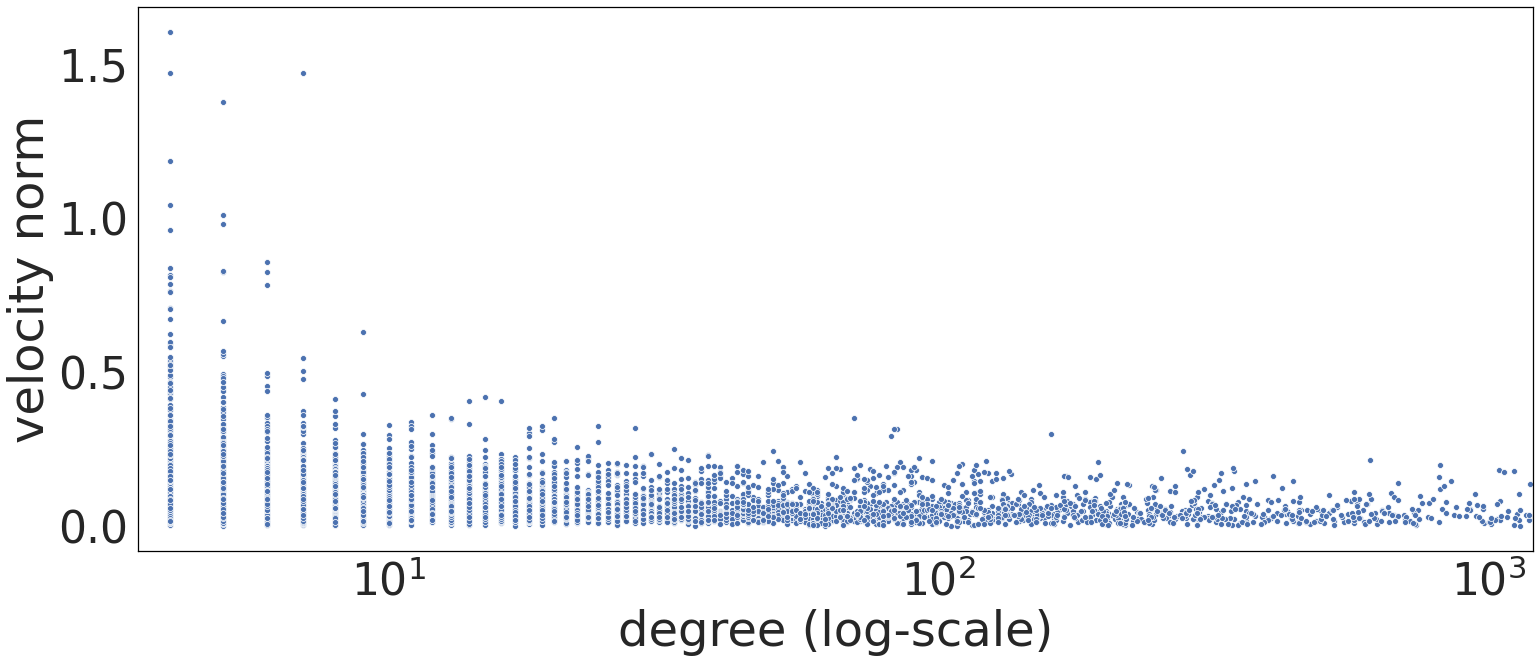}
\caption{\label{fig:degreeAndVelocityNorm} Scatter plot of velocity norms v.s. node degrees on ICEWS05-15. Each point denotes an entity.}
\end{figure}

\paragraph{Intrinsic hierarchical structures of temporal KGs} To illustrate geometric, especially the hierarchical, structures of temporal KGs, we focus on the Poincar\'e ball model with a dimension of $20$ and plot the geodesic distance $d_{\mathcal M}(\cdot, \mathbf 0)$ of learned entity embeddings to the origin of the Poincar\' e ball 
versus the degree of each entity in Figure \ref{fig:degreeAndDistanceOrigin}. Note that the distance is averaged over all timestamps since entity embeddings are time-dependent. We observe that entities with high degrees, which means they got involved in lots of facts, are generally located close to the origin. This makes sense because these entities often lie in the top hierarchical levels. And thus, they should stand close to the root. Under the same settings, we plot the velocity norm of each entity versus the entity degree in Figure \ref{fig:degreeAndVelocityNorm}.  Similarly, we see that entities with high degrees have a small velocity norm to stay near the origin of the manifold.

\begin{figure}
\begin{subfigure}{.24\textwidth}
 \centering
   \includegraphics[width=.9\linewidth]{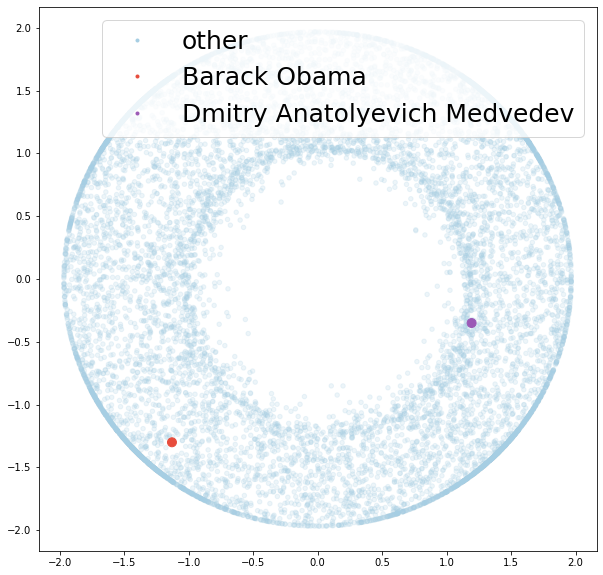}
\end{subfigure}%
\begin{subfigure}{.24\textwidth}
 \centering
  \includegraphics[width=.9\linewidth]{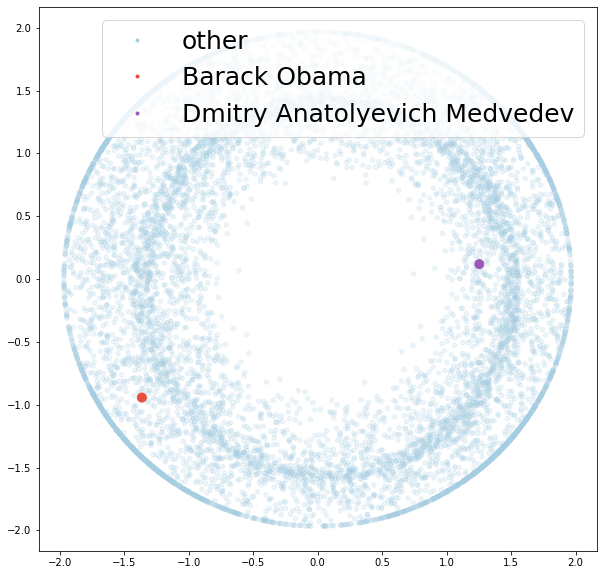}
\end{subfigure}%
\caption{\label{fig:Tendency_Obama_Medvedev} Learned two-dimensional hyperbolic entity embeddings of ICEWS05-15 on the first timestamp 2005-01-01 (left) and the last timestamp 2015-12-31 (right).}
\end{figure}

\paragraph{Relative movements between a node pair}
Figure \ref{fig:Tendency_Obama_Medvedev} shows two-dimensional hyperbolic entity embeddings of the ICEWS05-15 dataset on two timestamps, 2005-01-01 and 2015-12-31. Specifically, we highlight a former US president (in orange) and a former prime minister of Russia (in purple). We found that the interaction between these two entities decreased between 2005 and 2015, as shown in Figure \ref{fig: Interaction between country leaders} in the appendix. Accordingly, we observe that the embeddings of these two entities were moving away from each other. More examples of learned embeddings are relegated to Appendix F. 

\section{Conclusion}
In this paper, we propose an embedding approach for temporal knowledge graphs on a product of Riemannian manifolds with heterogeneous curvatures.  To capture the temporal evolution of temporal KGs, we use velocity vectors defined in tangent spaces to learn time-dependent entity representations. We show that our model significantly outperforms its Euclidean counterpart and other state-of-the-art approaches on three benchmark datasets of temporal KGs, which demonstrates the significance of geometrical spaces for the temporal knowledge graph completion task.

\section*{Acknowledgement}
The authors acknowledge support by the German Federal
Ministry for Education and Research (BMBF), funding project
MLWin (grant 01IS18050).

\bibliographystyle{acl_natbib}
\bibliography{Mixed_Curvature_Representations_for_Temporal_Knowledge_Graph_Completion}
\vspace{5mm}
\appendix
\section*{Appendices}
\vspace{2mm}
\section{Graph Curvature Estimation Algorithm}
We use Algorithm 1 to estimate the sectional curvatures of a dataset developed  by \citet{bachmann2019constant}.

\section{Implementation Details of Baselines} 
Note that the embedding dimension for each (baseline, dataset) pair
matches the number of parameters of our models with an embedding dimension of $100$. We use Table \ref{tab: Number of model parameters} and \ref{tab:dataset statistics} to compute the rank for each (baselines, dataset) pair. Besides, for fairness of results, we use the datasets augmented with reciprocal relations to train all baseline models.

\paragraph{Static knowledge graph embedding models}  
We use TransE \citep{bordes2013translating}, DistMult \citep{yang2014embedding}, and ComplEx \citep{trouillon2016complex}  as static baselines, where we compress temporal knowledge graphs into a static, cumulative graph by ignoring the time information. We use the cross-entropy loss and Adam optimizer with a batch size of 128 to train the static baselines. Besides, we use uniform sampling to initialize the embeddings of entities and predicates. Other hyperparameters  of the above baselines are shown in Table \ref{tab:Hyperparameter configurations of static baselines}.

\begin{table}[htbp]
  \begin{center}
    \caption{Number of parameters for each model considered when using reciprocal relations: $d$ represent the dimension of embeddings.}
    \label{tab: Number of model parameters}
    \begin{tabular}{cc}
      \toprule % <-- Toprule here
      Model & \# Parameters \\ 
      \midrule
      ComplEx &  $(2|\mathcal E| + 4|\mathcal P|)\cdot d$\\
      TransE & $(|\mathcal E| + 2|\mathcal P|)\cdot d$\\
      DistMult & $(|\mathcal E| + 2|\mathcal P|)\cdot d$ \\
      TComplEx &  $(2|\mathcal E| + 4|\mathcal P|+ 2|\mathcal T|)\cdot d$\\
      TTransE &  $(|\mathcal E| + 2|\mathcal P| + |\mathcal T|)\cdot d$\\
      TDistMult  & $(|\mathcal E| + 2|\mathcal P| + |\mathcal T|)\cdot d$\\
      HyTE & $(|\mathcal E| + 2|\mathcal P| + |\mathcal T|)\cdot d$ \\
      DyERNIE &  $2(|\mathcal E| + 2|\mathcal P|)\cdot d + 2 |\mathcal E|$\\
      \bottomrule % <-- Bottomrule here
    \end{tabular}
%}
  \end{center}
\end{table}

\paragraph{Temporal knowledge graph embedding models} 
We compare our model's performance with several state-of-the-art temporal knowledge graph embedding methods, including TTransE \citep{leblay2018deriving}, TDistMult/TComplEx \citep{ma2018embedding},
and HyTE \citep{dasgupta2018hyte}. We use the ADAM optimizer \citep{kingma2014adam} and the cross-entropy loss to train the temporal KG models. We set learning rate = $0.001$, negative samples pro fact = $500$, number of epochs = $500$ , batch size = $256$, and validate them every $50$ epochs to select the model giving the best validation MRR. For the GDELT dataset, we use a similar setting but with negative samples pro fact = $50$ due to the large size of the dataset. The embedding dimensions of the above dynamic baselines on each dataset are shown in Table \ref{tab:Embedding dimensions of dynamic baselines}.

\begin{table}[htbp]
 \caption{Hyperparameter settings of static baselines.}
    \label{tab:Hyperparameter configurations of static baselines}
\resizebox{\linewidth}{!}{
\begin{tabular}{l|lll}
    \toprule
    Model&\multicolumn{1}{c}{TransE}&\multicolumn{1}{c}{DistMult}&\multicolumn{1}{c}{ComplEx}\\ 
    \midrule
    Embedding dimension\\
    \quad ICEWS14 & 202 & 202 & 101 \\
    \quad ICEWS05-15 & 202 & 202 & 101 \\
    \quad GDELT & 202 & 202 &  101\\
    Negative Sampling & 253 & 657 & 1529\\
    Learning rate & 3e-4 & 0.16 & 0.18\\
    \bottomrule
\end{tabular}
}
\end{table}

\begin{table} 
 \caption{Embedding dimensions of dynamic baselines.}
    \label{tab:Embedding dimensions of dynamic baselines}
\resizebox{\linewidth}{!}{
\begin{tabular}{l|llll}
    \toprule
    Model&\multicolumn{1}{c}{TTransE}&\multicolumn{1}{c}{TDistMult}&\multicolumn{1}{c}{TComplEx}&HyTE\\ 
    \midrule
    Embedding dimension\\
    \quad ICEWS14 & 193  & 193 & 96 & 193\\
    \quad ICEWS05-15 & 148 & 148 & 74 & 148\\
    \quad GDELT & 151 & 151 & 76  & 151\\
    \bottomrule
\end{tabular}
}
\end{table}

\section{Datasets}
Dataset statistics are described in Table \ref{tab:dataset statistics}. Since the timestamps in
the ICEWS dataset are dates rather than numbers, we sort them chronologically and encode them into consecutive numbers.

\section{Evaluation metrics}
Let $\psi_{e_s}$ and $\psi_{e_o}$ represent the rank for $e_s$ and $e_o$ of the two queries, respectively. We evaluate our models using standard metrics across the link prediction literature: \textit{mean reciprocal rank (MRR)}: $\frac{1}{2\cdot |\mathcal G_{test}|} \sum_{q \in \mathcal G_{test}}(\frac{1}{\psi_{e_s}} + \frac{1}{\psi_{e_o}})$ and \textit{Hits}$@k(k \in \{1,3,10\})$: the percentage of times that the true entity candidate appears in the top $k$ of ranked  candidates. 

\section{Implementation Details of DyERNIE}

\paragraph{Signature search}
On the ICEWS subsets, we try all manifold combinations with the number of components of $\{1,2,3\}$. Due to the large size of data samples on the GDELT dataset, we only try manifold combinations with the number of components of $\{1,2\}$. Specifically, the candidates are $\{\mathbb P^n, \mathbb S^n,  \mathbb E^n \}$ for single manifolds, $\{\mathbb P^{n_i} \times \mathbb S^{n_i}, \mathbb P^{n_i} \times \mathbb P^{n_i},  \mathbb S^{n_i} \times \mathbb S^{n_i}, \mathbb P^{n_i} \times \mathbb E^{n_i}, \mathbb S^{n_i} \times \mathbb E^{n_i}\}$ for a product of two component manifolds, and $\{\mathbb P^{n_i} \times \mathbb P^{n_i} \times \mathbb P^{n_i},  \mathbb P^{n_i} \times \mathbb S^{n_i} \times \mathbb E^{n_i}, \mathbb S^{n_i} \times \mathbb S^{n_i} \times \mathbb S^{n_i}, \mathbb P^{n_i}, \times \mathbb P^{n_i} \times \mathbb S^{n_i}, \mathbb P^{n_i} \times \mathbb S^{n_i} \times \mathbb S^{n_i}, \mathbb P^{n_i} \times \mathbb P^{n_i} \times \mathbb E^{n_i}, \mathbb S^{n_i} \times \mathbb S^{n_i} \times \mathbb E^{n_i}\}$ for a product of three component manifold. 
For each combination, we use the Ax-framework\footnote{https://ax.dev} to optimize the assignment of dimensions to each component manifold and the curvatures. The assignment of the best-performing models are shown in 
Table \ref{tab:Hyperparameter configurations for best-performing models on the ICEWS14 dataset}, \ref{tab:Hyperparameter configurations for best-performing models on the ICEWS05-15 dataset}, and \ref{tab:Hyperparameter configurations for best-performing models on the GDELT dataset}. We report the best results on each dataset in Table 1 in the main body. 

\paragraph{Hyperparameter configurations for best-performing models}
We select the loss function from binary cross-entropy (BCE), margin ranking loss, and cross-entropy (CE). BCE and CE give a similar performance and outperform the margin ranking loss. However, when using the BCE loss, we could use a large learning rate ($lr>10$) to speed up the training procedure. In contrast, models with the CE loss incline overfitting by large learning rates. Given the BCE loss, we found the learning rate of $50$ works the best for all model configurations. 
Furthermore, increasing negative samples can improve the performance to some extent, while this impact is weakening gradually as the number of negative samples become larger. 
However, the number of negative samples largely affect the runtime of the training procedure. We empirically found that  the negative sample number of $50$ is a good compromise between the model performance and the training speed. Besides, there is no statistically significant difference in the model performance when using different optimizers, such as Riemannian Adam (RADAM) and Riemannian stochastic gradient descent (RSGD).  Thus, for the model's simplicity, we decide to use RSGD. 

\paragraph{Average runtime for each approach $\&$ Number of parameters in each model}
Table \ref{tab: Average runtime and number of parameters for each approach} shows the number of parameters and the average runtime for each model.

\section{Visualization}
\begin{figure*}[htbp]
    \centering
    \includegraphics[width=.8\textwidth]{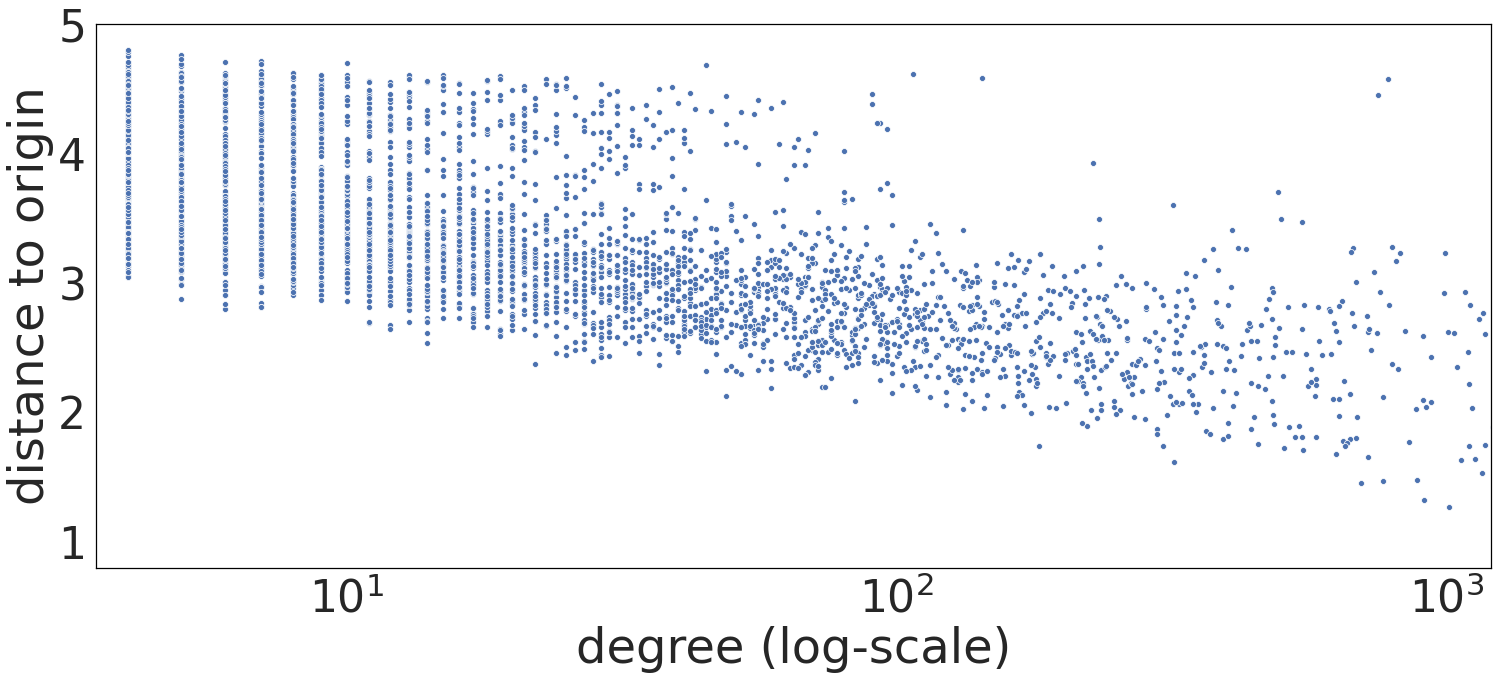}
    \includegraphics[width=.8\textwidth]{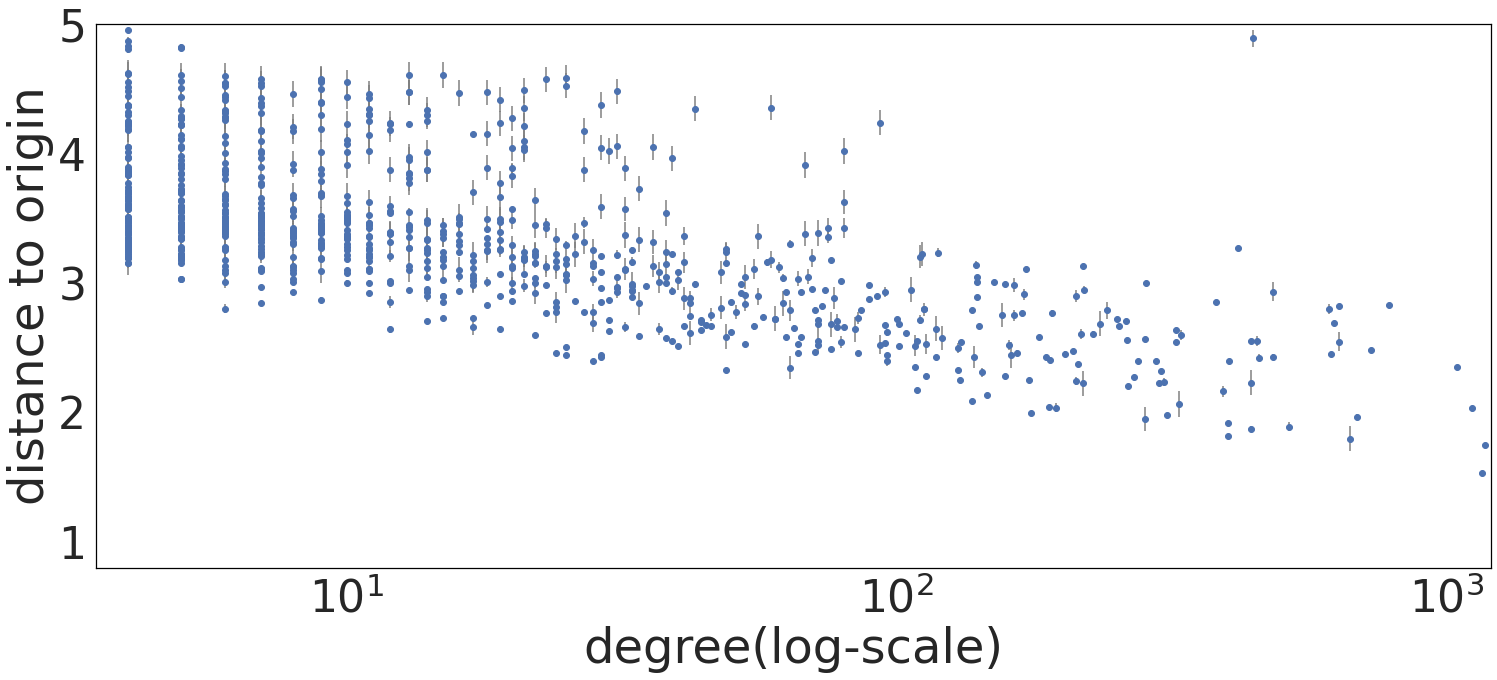}
    \caption{Each point in the upper plot represents an entity whose x-coordinate gives their degree accumulated over all timestamps and y-coordinate gives their distance to the origin averaged over all timestamps. The plot clearly shows the tendency that entities with high degrees are more likely to lie close to the origin. The bottom plot shows the same content but with a sampling of 20\% points. The gray bar around each point shows the variance of the distance over all timestamps.}
    \label{fig:dist2center-degree}
\end{figure*}

We plot the geodesic distance $d_{\mathcal M}(\mathbf e_j, \mathbf 0)$ of learned entity embeddings with a dimension of $20$ to the manifold's origin versus the degree of each entity in Figure \ref{fig:dist2center-degree}, where $d_{\mathcal M}(\mathbf e_j, \mathbf 0)$ is averaged over all timestamps since $\mathbf e_j$ is time-dependent. Also, the degree of each entity is accumulated over all timestamps. Each point in the upper plot represents an entity where the x-coordinate gives their degree, and the y-coordinate gives their average distance to the origin. The plot clearly shows the tendency that entities with high degrees are more likely to lie close to the origin. The bottom plot shows the same content but with a sampling of 20\% points. The gray bar around each point shows the variance of the distance between the entity embedding and the origin over time.

\begin{figure*}[htbp]
    \centering
    \begin{subfigure}[htbp]{0.24\textwidth}
        \centering
        \includegraphics[width=.99\textwidth]{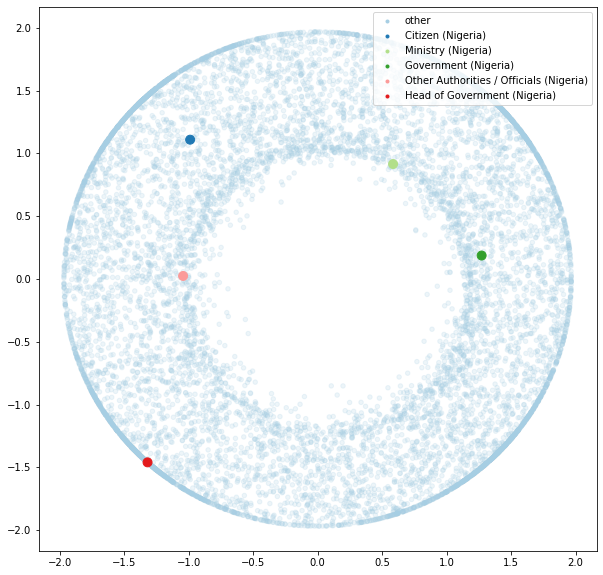}
    \end{subfigure}
    \begin{subfigure}[htbp]{0.24\textwidth}
        \centering
        \includegraphics[width=.99\textwidth]{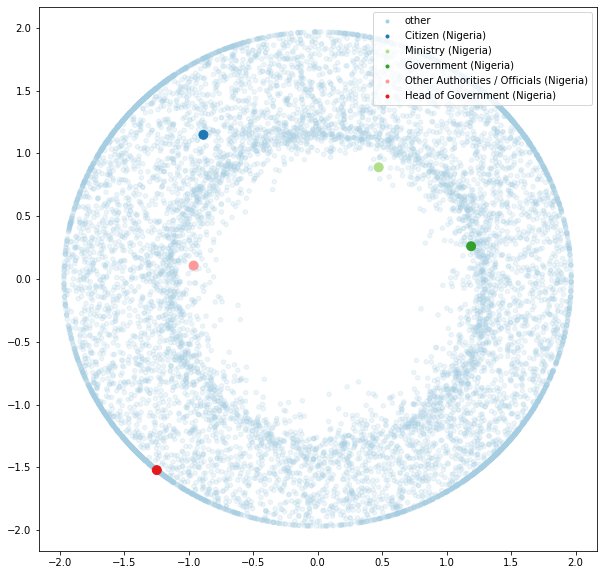}
    \end{subfigure}
    \begin{subfigure}[htbp]{0.24\textwidth}
        \centering
        \includegraphics[width=0.99\textwidth]{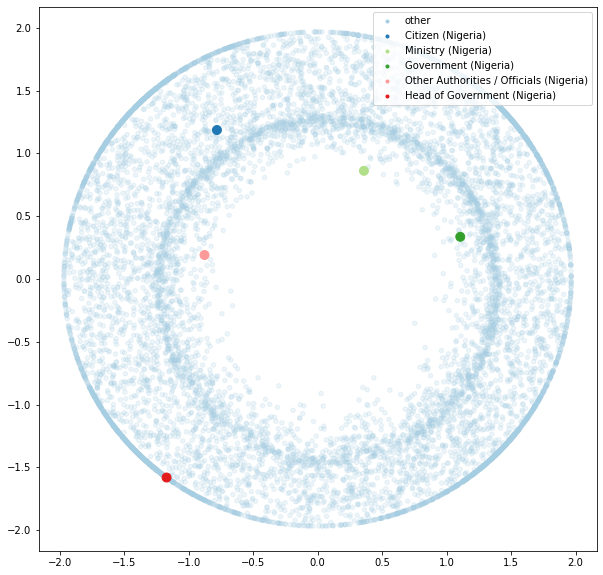}
    \end{subfigure}
    \begin{subfigure}[htbp]{0.24\textwidth}
        \centering
        \includegraphics[width=0.99\textwidth]{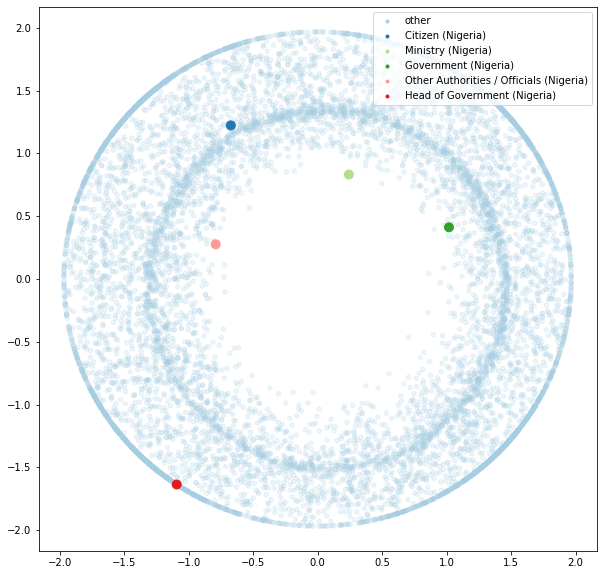}
    \end{subfigure}
    \caption{Evolution of entity embeddings over time.  We highlight Nigerian citizens, the Nigerian government, the head of Nigerian government, other authorities in Nigeria, and Nigerian minister.}
    \label{fig: Evolution of embedding of entities over time. }
\end{figure*}

\begin{figure*}[htbp]
    \centering
    \begin{subfigure}[t!]{0.24\textwidth}
        \centering
        \captionsetup{justification=centering}
        \includegraphics[width=.99\textwidth]{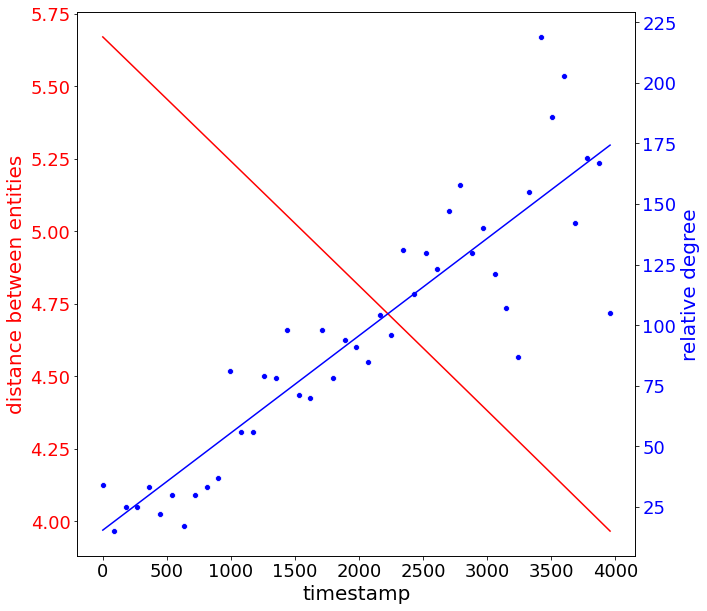}
        \caption{Citizen (Nigeria)\newline Government (Nigeria)}
    \end{subfigure}
    \begin{subfigure}[t!]{0.24\textwidth}
        \centering
        \captionsetup{justification=centering}
        \includegraphics[width=.99\textwidth]{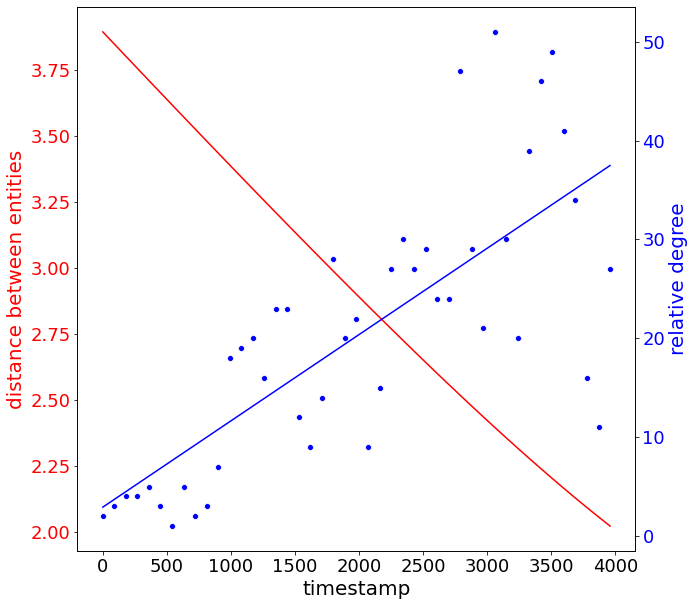}
        \caption{Citizen (Nigeria)\newline Ministry (Nigeria)}
    \end{subfigure}
    \begin{subfigure}[t!]{0.24\textwidth}
        \centering
        \captionsetup{justification=centering}
        \includegraphics[width=0.99\textwidth]{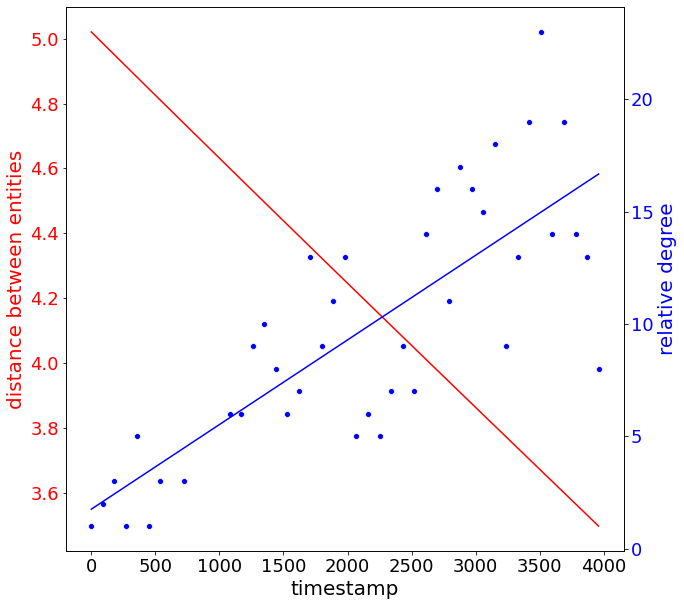}
        \caption{Government (Nigeria)\newline Other Authorities (Nigeria)}
    \end{subfigure}
    \begin{subfigure}[t!]{0.24\textwidth}
        \centering
        \captionsetup{justification=centering}
        \includegraphics[width=0.99\textwidth]{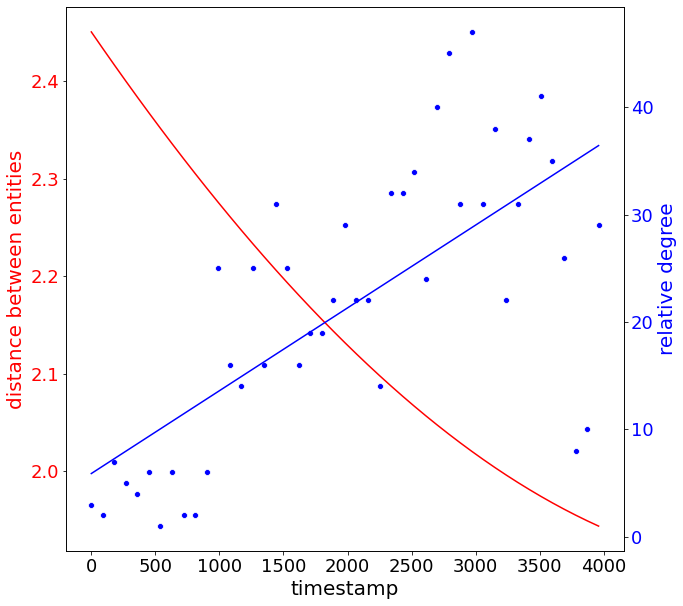}
        \caption{Government (Nigeria)\newline Ministry (Nigeria)}
    \end{subfigure}
    \caption{Interaction between Nigerian entities. Subtitles show the names of the given entity pair. Red lines give the geodesic distance between two entities. Blue dots represent the number of interactions between two entities (relative degree) at each timestamp, and blue lines are regression of the relative degree between two entities over time.}
      \label{fig: Interaction between Nigerian stuff.}
\end{figure*}

\begin{figure*}[htbp]
    \centering
    \begin{subfigure}[t!]{.46\textwidth}
        \centering
        \includegraphics[width=.99\textwidth]{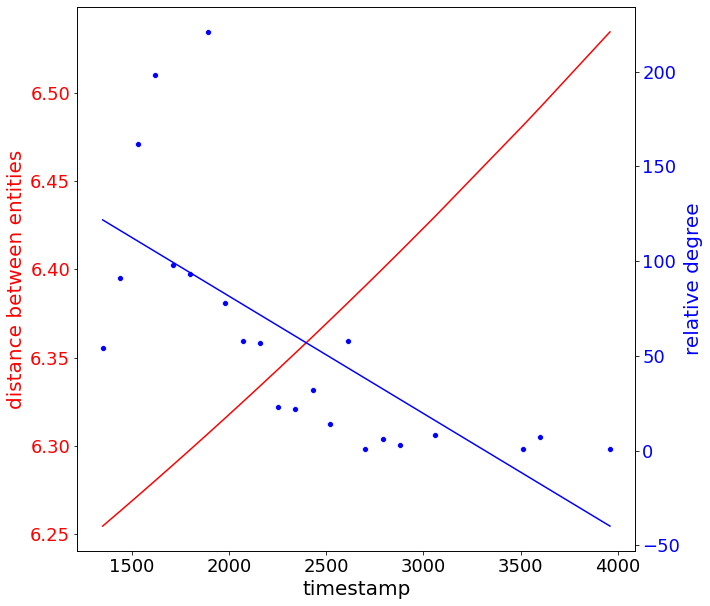}
    \end{subfigure}
       
    \caption{Interaction between Barack Obama and Dmitry Anatolyevich Medvedev.  Red lines give the geodesic distance between two entities. Blue dots represent the number of interactions between two entities (relative degree) at each timestamp, and blue lines are regression of the relative degree between two entities over time.}
    \label{fig: Interaction between country leaders}
\end{figure*}

Figure \ref{fig: Evolution of embedding of entities over time. } shows two-dimensional hyperbolic entity embeddings of the ICEWS05-15 dataset on four timestamps. We highlight some entities to show the relative movements between them. The number of interactions between the selected entities are depicted in Figure \ref{fig: Interaction between Nigerian stuff.} and \ref{fig: Interaction between country leaders}, which evolves over time. Specifically, we highlight Nigerian citizens, the Nigerian government, head of the Nigerian government, other authorities in Nigeria, and Nigerian minister in the first row of subplots. %Furthermore, we show the relative movements between the entity embeddings of Barack Obama and Dmitry Anatolyevich Medvedev in the second row of subplots. 
We can see that two entities were getting close in the Poincare disc if the number of interactions between them increases.

\section{Additional Ablation Study}
\label{app: additional ablation study}
To assess the contribution of the dynamic part of entity embeddings, we remove the dynamic part and run the model variant on static knowledge graphs. Specifically, we compress ICEWS05-15 into a static, cumulative graph by ignoring the time information. As shown in Table \ref{tab: time representation variantions},  the performance degrades significantly if the entity embeddings only have the static part. For example, on the ICEWS0515 dataset, the Hits@1 metric of DyERNIE-Sgl in the static case is less than half of that in the dynamic case, clearly showing the gain from the dynamism. 

\begin{table}[htbp]
  \begin{center}
    \caption{Filtered MRR for dynamic/static entity representations with $dim = 20$ on ICEWS05-15. Note that we run the static model variant on static ICEWS05-15.}
    \label{tab: time representation variantions}
    \resizebox{0.4\textwidth}{!}{
    \begin{tabular}{lcccc}
      \toprule % <-- Toprule here
  	Entity Representations & MRR & Hits@1 & Hits@3 & Hits@10\\
  	 \midrule % <-- Midrule here
  	With dynamic part & 63.8 & 55.9 & 67.9 & 78.7 \\
  	  \midrule % <-- Midrule here
	Without dynamic part & 38.6 & 28.3 & 42.8 & 59.2\\ 
      \bottomrule % <-- Bottomrule here
    \end{tabular}
}
  \end{center}
\end{table}

\begin{algorithm*}[ht!]
\SetAlgoLined
 \SetKwInOut{Input}{Input}
 \SetKwInOut{Output}{Output}
 \Input{Number of iterations $n_{iter}$, number of timestamps $n_{time}$, Graph Slices $\{G_i\}_{i=1}^{n_{time}}$ of a temporal knowledge graph, Neighbor dictionary $\mathcal N$.}
    \Output{$\{K_i\}_{i=1}^{n_{time}}$}
    \BlankLine
\For { $i=1$ \textbf{to} $n_{time}$}{
	\For {$m \in G_i$}{
	 	\For{$j=1$ \textbf{to} $n_{iter}$}{
	  $b, c \sim \mathcal U(\mathcal N(m))$ and $a \sim \mathcal U(G_i \backslash \{m\})$ \\
	   $\psi_j(m,b,c,a) = \frac{1}{2d_{G_i}(a,m)}\left(2d_{G_i}^2(a,m) + d_{G_i}^2(b,c)/4 - d_{G_i}^2(a,b)/2 + d_{G_i}^2(a,c)/2\right)$
	}
	 $\psi_i(m) = \sum_{j=1}^{n_{iter}}\psi_j(m, b, c, a)$
	 }
	 $K_i = \sum_{m \in G_i} \psi_i(m)$
		}
    \BlankLine
 \caption{Curvature Estimation}
 \label{alg: Curvature Estimation}
\end{algorithm*}

\begin{table*}[htbp]
  \begin{center}
    \caption{Exponential and logarithmic maps in Poincar\' e ball and projected hypersphere.}
    \label{tab: Exponential and logarithmic maps}
   % \resizebox{80mm}{8mm}{
    \begin{tabular}{cc}
      \toprule % <-- Toprule here
      trigonometric functions & $\tan_K (\cdot) = \tan (\cdot)\,$ if $K > 0$; $\;\tanh (\cdot)\,$ if $K < 0$\\
      Exponential map &  $\exp^K_{\mathbf x}(\mathbf v) = \mathbf x \oplus (\tan_K(\frac{\sqrt{|K|}\lambda^K_{\mathbf x}||\mathbf v||_2}{2})\frac{\mathbf v}{\sqrt{K} ||v||_2})$\\
      Logarithmic map &  $\log^K_{\mathbf x}(\mathbf v) = \frac{2}{\sqrt{|K|}\lambda^K_{\mathbf x}} \tan^{-1}_K(\sqrt{|K|}||-\mathbf x \oplus_K \mathbf v||_2)\frac{-\mathbf x \oplus_K \mathbf v}{||\mathbf x \oplus_K \mathbf v||_2}$\\
      \bottomrule % <-- Bottomrule here
    \end{tabular}
%}
  \end{center}
\end{table*}
 
\begin{table*}
 \caption{Hyperparameter configurations for best-performing models on the ICEWS14 dataset.}
    \label{tab:Hyperparameter configurations for best-performing models on the ICEWS14 dataset}
\resizebox{\textwidth}{!}{
\begin{tabular}{lllll|llll|llll}
    \toprule
    Model&\multicolumn{4}{c|}{DyERNIE-Sgl}&\multicolumn{4}{c|}{DyERNIE-Prod}&\multicolumn{4}{c}{DyERNIE-Euclid}\\ 
    \midrule
     Embedding size &10&20&40&100&10&20&40&100&10&20&40&100\\ 
	Curvature \\
   	\quad Component A& -0.172 & -0.171 & -0.171 & -0.170  &-0.044 & -0.114  & -0.177 & -0.346 & 0 & 0 & 0 & 0\\
    \quad Component B & - & - & - & - &-0.128&-0.286&-0.281&-0.137& - & - & - & - \\
    \quad Component C & - & - & - & - &-0.371& -0.422 & -0.470 &-0.855& - & - & - & - \\
    Dimension scale\\
    \quad Component A & 10 & 20 & 40 & 100 & 3 & 14 & 20 & 20 & 10 & 20 & 40 & 100\\
    \quad Component B & - & - & - & - & 1 & 4 & 8 & 21 & - & - & - & -  \\
    \quad Component C & - & - & - & - & 6 & 2 & 12 & 59  & - & - & - & - \\
    \bottomrule
\end{tabular}
}
\end{table*}

\begin{table*}
 \caption{Hyperparameter configurations for best-performing models on the ICEWS05-15 dataset.}
    \label{tab:Hyperparameter configurations for best-performing models on the ICEWS05-15 dataset}
\resizebox{\textwidth}{!}{
\begin{tabular}{lllll|llll|llll}
    \toprule
    Model&\multicolumn{4}{c|}{DyERNIE-Sgl}&\multicolumn{4}{c|}{DyERNIE-Prod}&\multicolumn{4}{c}{DyERNIE-Euclid}\\ 
    \midrule
     Embedding size &10&20&40&100&10&20&40&100&10&20&40&100\\ 
	Curvature \\
   	\quad Component A& -0.180 & -0.181 & -0.179 & -0.178  & -0.102  & -0.122  & -0.298 & -0.453 & 0 & 0 & 0 & 0\\
    \quad Component B & - & - & - & - & -0.135  &  -0.163 & -1.243 & -0.216 & - & - & - & - \\
    \quad Component C & - & - & - & - & -0.214  & -0.191 & -1.819 &  -0.938 & - & - & - & - \\
    Dimension scale\\
    \quad Component A & 10 & 20 & 40 & 100 & 7 & 10 & 31 & 32 & 10 & 20 & 40 & 100\\
    \quad Component B & - & - & - & - & 2 & 8 & 5 & 52 & - & - & - & -  \\
    \quad Component C & - & - & - & - & 1 & 2 & 4 & 16  & - & - & - & - \\
    \bottomrule
\end{tabular}
}
\end{table*}

\begin{table*} 
 \caption{Hyperparameter configurations for best-performing models on the GDELT dataset.}
    \label{tab:Hyperparameter configurations for best-performing models on the GDELT dataset}
\resizebox{\textwidth}{!}{
\begin{tabular}{lllll|llll|llll}
    \toprule
    Model&\multicolumn{4}{c|}{DyERNIE-Sgl}&\multicolumn{4}{c|}{DyERNIE-Prod}&\multicolumn{4}{c}{DyERNIE-Euclid}\\ 
    \midrule
     Embedding size &10&20&40&100&10&20&40&100&10&20&40&100\\ 
	Curvature \\
   	\quad Component A& 0.279 & 0.336 & 0.259 & 0.197  & 0.213  & 0.241  & 0.202 & 0.342 & 0 & 0 & 0 & 0\\
    \quad Component B & - & - & - & - & 0.291  & 0.336 & 0.291 & 0.336 & - & - & - & - \\
    Dimension scale\\
    \quad Component A & 10 & 20 & 40 & 100 & 8 & 8 & 10 & 68 & 10 & 20 & 40 & 100\\
    \quad Component B & - & - & - & - & 2 & 12 & 30 & 32 & - & - & - & -  \\
    \bottomrule
\end{tabular}
}
\end{table*}

\begin{table*}[htbp]
  \begin{center}
    \caption{Datasets Statistics }
    \label{tab:dataset statistics}
   % \resizebox{80mm}{8mm}{
    \begin{tabular}{cccccccc}
      \toprule % <-- Toprule here
      Dataset Name & $|\mathcal E|$  & $|\mathcal P|$ & $|\mathcal T|$ & $|\mathcal G|$ & |$train$| & |$validation$| & |$test$|\\
      \midrule % <-- Midrule here
      ICEWS14 & 7,128 & 230 & 365 & 90,730 & 72,826 & 8,941 & 8,963\\
      ICEWS05-15 & 10,488 & 251 & 4,017 & 479,329 & 386,962 & 46,275 & 46,092 \\
      GDELT & 7,691 & 240 & 2,975 & 2,278,405 & 1,734,399 & 238,765 & 305,241\\ 
      \bottomrule % <-- Bottomrule here
    \end{tabular}
%}
  \end{center}
\end{table*}

\begin{table*}[ht!] 
    \caption{Average runtime and parameter number for each approach: runtime is in seconds.}
    \label{tab: Average runtime and number of parameters for each approach}
    \begin{center}
      \resizebox{\textwidth}{!}{
    \begin{tabular}{l|l|ccc|ccc|ccc} 
      \toprule % <-- Toprule here
     \multicolumn{2}{l}{Datasets} & \multicolumn{3}{|c}{\textbf{ICEWS14}} &  \multicolumn{3}{|c}{\textbf{ICEWS05-15}} & \multicolumn{3}{|c}{\textbf{GDELT}}\\
      \midrule % <-- Midrule here
      Rank $(d)$ & Model & Manifold & Runtime & Parameters & Manifold & Runtime & Parameters & Manifold & Runtime & Parameters\\
       \midrule % <-- Midrule here
        & TransE & & 3,800  & 1,531,856 & &15,200 & 2,218,976 &  & 85,600 & 1,649,582 \\
       100 & DistMult & $\mathbb E$ & 9,900 & 1,531,856 & $\mathbb E$ & 31,500 & 2,218,976 & $\mathbb E$ & 132,700 & 1,649,582\\
        & ComplEx &  &  4,300 & 1,531,856 & &  14.100 & 2,218,976 & & 76,000 & 1,649,582\\
      \midrule % <-- Midrule here
       & TTransE & & 55,000 & 1,531,856 & & 430,000&  2,218,976 & & 1,500,000 & 1,649,582 \\
       & TDistMult & & 85,000 & 1,531,856 & & 680,000 & 2,218,976 & & 2,040,000 &  1,649,582 \\
      100 & TComplEx & $\mathbb E$ & 65,000 &  1,531,856 & $\mathbb E$ & 520,000 & 2,218,976 &  $\mathbb E$  & 1,500,000 & 1,649,582 \\
       & HyTE & & 45,000 & 1,531,856 &  & 360,000 & 2,218,976 & & 1,100,000 & 1,649,582\\
        \midrule % <-- Midrule here
         & DyERNIE-Prod & $\mathbb P^3$ & 44,500 & 1,531,856 & $\mathbb P^3$ & 343,800 & 2,218,900 &  $\mathbb S^2$ & 1,2	59,400 & 1,649,582 \\       
          100 & DyERNIE-Sgl & $\mathbb P$ & 42,000 &  1,531,856 &  $\mathbb P$ & 341,900 & 2,218,976 & $\mathbb S$ & 1,208,300 & 1,649,582 \\
        & DyERNIE-Euclid & $\mathbb E$ & 19,000 & 1,531,856 & $\mathbb E$ & 38,000 &  2,218,976 & $\mathbb E$ & 388,800 & 1,649,582\\
           \midrule % <-- Midrule here
         & DyERNIE-Prod &  $\mathbb P^3$  & 35,500 & 621,296 & $\mathbb P^3$ & 229,500 & 900,176 &  $\mathbb S^2$ & 800,000 & 669,062\\
         40 & DyERNIE-Sgl  & $\mathbb P$  & 32,000 & 621,296 &$\mathbb P$  & 225,000 & 900,176 & $\mathbb S $ &740,000 & 669,062\\
          & DyERNIE-Euclid & $\mathbb E$ &11,000 & 621,296 & $\mathbb E$ & 25,000 & 900,176 & $\mathbb E$ & 262,000 & 669,062\\
            \midrule % <-- Midrule here
       & DyERNIE-Prod & $\mathbb P^3$  & 32,500 & 317,776 & $\mathbb P^3$ & 225,000 & 460,576 & $\mathbb S^2$ & 700,000 & 342,222\\
           20 & DyERNIE-Sgl  & $\mathbb P$  & 31,500 & 317,776 & $\mathbb P$  & 220,000 & 460,576  & $\mathbb S$ & 676,000 &  342,222\\  
              & DyERNIE-Euclid  & $\mathbb E$ & 9,500 &  317,776 & $\mathbb E$ & 22,000 & 460,576 & $\mathbb E$ & 240,000 & 342,222\\
          \midrule % <-- Midrule here
        & DyERNIE-Prod & $\mathbb P^3$ &  20,500 & 166,016  & $\mathbb P^3$ & 165,000 &  240,776 & $\mathbb S^2$ & 420,000 & 178,802 \\
           10 & DyERNIE-Sgl  & $\mathbb P$ & 20,500 & 166,016 & $\mathbb P$  & 150,000 &  240,776 & $\mathbb S$ & 400,000 & 178,802 \\ 
           & DyERNIE-Euclid & $\mathbb E$ & 6,500 & 166,016 & $\mathbb E$ & 15,000 & 240,776 & $\mathbb E$ & 180,000 & 178,802\\
      \bottomrule % <-- Bottomrule here
    \end{tabular} }
    \end{center}
	\end{table*}
  
\end{document}